\pdfoutput=1
\documentclass[11pt]{article}

\usepackage[]{EMNLP2023}

\usepackage{times}
\usepackage{latexsym}
\usepackage{booktabs}
\usepackage{multirow}
\usepackage[utf8]{inputenc}
\usepackage{graphicx}
\usepackage{amssymb}
\usepackage{pifont}
\usepackage{listings} % Load the listings package
\usepackage{xcolor} % Optional: for color customization

\definecolor{cadmiumgreen}{rgb}{0.0, 0.42, 0.24}
\definecolor{celestialblue}{rgb}{0.29, 0.59, 0.82}
\newcommand{\myorange}[1]{\textcolor{orange}{#1}}
\newcommand{\mygreen}[1]{\textcolor{cadmiumgreen}{#1}}
\newcommand{\myblue}[1]{\textcolor{celestialblue}{#1}}
\usepackage[T1]{fontenc}
\usepackage[utf8]{inputenc}

\usepackage{microtype}

\usepackage{inconsolata}

\title{Exploring Explainable Automated Student Answer Assessment with ChatGPT}
\title{Distilling ChatGPT for Explainable Automated Student Answer Assessment}
\author{Jiazheng Li$^1$, Lin Gui$^1$, Yuxiang Zhou$^1$, David West$^2$, Cesare Aloisi$^2$ and Yulan He$^{1,3}$\\
  $^1$Department of Informatics, King's College London, UK\\
  $^2$AQA, UK~~~~~~$^3$The Alan Turing Institute, UK\\
\texttt{\{jiazheng.li, lin.1.gui, yuxiang.zhou, yulan.he\}@kcl.ac.uk}\\
  \texttt{\{caloisi, dwest\}@aqa.org.uk}}
  
\begin{document}
\maketitle
\begin{abstract}
% Assessing student answers and providing valuable feedback is crucial for effective learning, but it can be a time-consuming task.
% Providing timely and informative feedback is crucial for automated student answer assessment.
Providing explainable and faithful feedback is crucial for automated student answer assessment.
%While recent research studies for explainable rationale generation are effective, such methods typically require an abundance of human annotations.
%Zero-shot prompting and few-shot prompting can potentially lead to the problem where the rationale and label may be inconsistent.
% Most recent research efforts directly determine the correctness of student answers through the classification paradigm.
% Various research studies either relied on manual annotation or introduced free-text rationale without matching them with the \textit{rubric} and \textit{key elements}, leading to \textit{inconsistency} in the rationales.
% While these approaches are empirically effective, they typically fail to provide explanations for the automated assessment process.
% Traditional methods of automating student answer assessment through text classification often suffer from issues such as lack of trustworthiness, transparency, and the ability to provide a rationale for the automated assessment process. These limitations hinder their usefulness in practice. 
%Traditional text classification methods used for automating student response assessment are often less trustworthy, lack of transparency, and able to provide a rationale for the automated assessment process, thus limiting their usefulness. 
In this paper, we introduce a novel framework that explores using ChatGPT, a cutting-edge large language model, for the concurrent tasks of student answer scoring and rationale generation.
% We introduce a novel framework that prompts ChatGPT to generate rationales with assessment scores and rationale refinement.
We identify the appropriate instructions by prompting ChatGPT with different templates to collect the rationales, where inconsistent rationales are refined to align with marking standards.
The refined ChatGPT outputs enable us to fine-tune a smaller language model that simultaneously assesses student answers and provides rationales.
% under both the zero-shot and few-shot settings. We introduce a critic module which automatically filters incorrect outputs from ChatGPT and utilizes the remaining ChtaGPT outputs as noisy labelled data to fine-tune a smaller language model, enabling it to perform student answer scoring and rationale generation. Moreover, by drawing multiple samples from ChatGPT outputs, we are able to compute predictive confidence scores, which in turn can be used to identify corrupted data and human label errors in the training set. 
%We model the traditional scoring classification task into a more comprehensive classification-oriented explanation generation problem. By fine-tuning smaller language models on ChatGPT-generated rationales, they can be distilled to generate more reasonable rationales while achieving higher assessment performance compared with ChatGPT. 
% Our experimental results demonstrate that despite being a few orders of magnitude smaller than ChatGPT, 
% Our experimental results demonstrate that the fine-tuned language model achieves better performance in student answer assessment. 
Extensive experiments on the benchmark dataset show that the proposed method improves the overall QWK score by $11\%$ compared to ChatGPT.
Furthermore, our thorough analysis and human evaluation demonstrate that the rationales generated by our proposed method are comparable to those of ChatGPT.
% Furthermore, it generates more detailed and comprehensible assessments than traditional text classification methods without further requirements for annotation. 
Our approach provides a viable solution to achieve explainable automated assessment in education\footnote{Code available at \href{https://github.com/lijiazheng99/aera}{https://github.com/lijiazheng99/aera.}}. %offering a promising approach to enhance educational assessment and feedback.
\end{abstract}

\section{Introduction}
\begin{figure}[t]
  \centering
  \includegraphics[width=\linewidth]{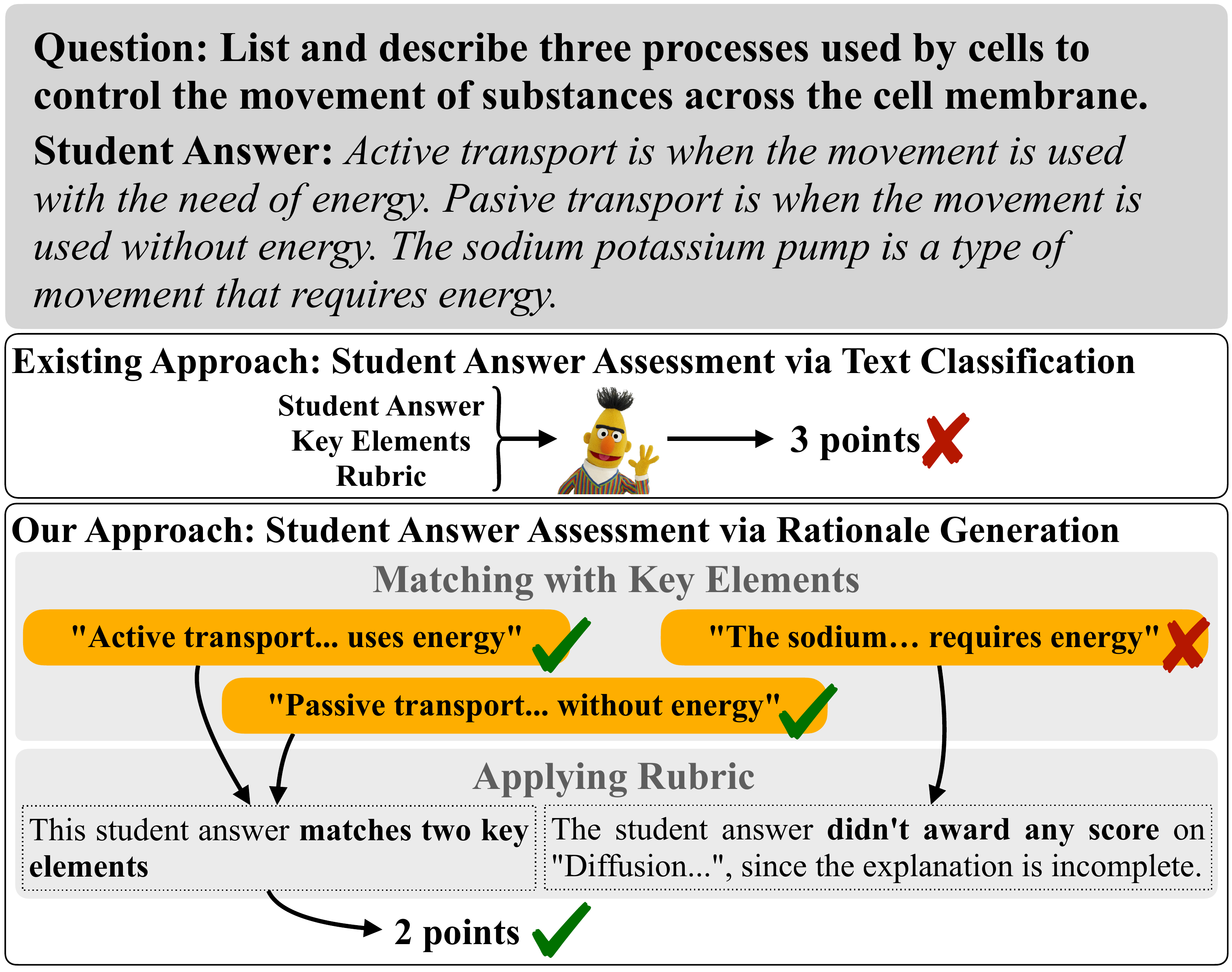}
  \caption{Classification-based automated answer assessment techniques often rely on black-box models, making the assessment process challenging to interpret. Incorporating rationale generation can significantly enhance the transparency of the assessment decisions.}
  \label{fig:sample}
\end{figure}
% What's the drawback of manual student answer assessment
% It can reflect the effectiveness of teachers' teaching methods and provide students with valuable feedback to identify their strengths and weaknesses \cite{nicol2006formative}. Timely and hattie2007power, nicol2006formative
Student answer assessment is a critical component of the education process. Prompt and insightful answer assessments can enhance students' learning experiences and support their academic growth \cite{nicol2006formative}. However, manually providing detailed feedback is time-consuming, and differences in assessment criteria among various evaluators can result in inconsistencies in the grading process \cite{weigle2002assessing}. %Therefore, there is a need for a more efficient and uniform method of student answer assessment in education.
% What's the drawback of classification-based student answer assessment

Various automated student answer assessment models have been proposed in recent years, mostly built on the Pre-trained Language Models (PLMs) \cite{devlin2018bert}, making the assessment process more efficient and consistent. These approaches \cite{sung2019improving, mayfield-black-2020-fine} tend to frame the assessment task as a classification problem, which involves training text classifiers to predict scores given student answers. However, as shown in Figure \ref{fig:sample}, the feedback provided in terms of scores is not sufficiently detailed for students to identify weaknesses in their answers. Besides, %, PLMs are black-box models. 
it is challenging for humans to interpret the classifiers' decision-making process, making classifiers' assessment results less trustworthy.
% However, when applied to student response assessment, these models still suffer from a lack of transparency and trustworthiness, as they tend to produce predictions without offering an explicit rationale behind their decisions. % \cite{riordan2017automated}. 
% This lack of transparency can be a significant drawback, as students and educators may not understand the reasoning behind the assessments, limiting the usefulness of the feedback. This issue prevents students and educators from understanding the model's thought process and limits its applicability in educational settings.

% Why rationale can help student answer assessment
% Researchers developed ways to
Researchers have advocated for generating rationales to enhance the interpretability of classifiers. These rationales are natural language explanations that substantiate model predictions \cite{Gurrapu2023RationalizationFE}. Often, such strategies necessitate rationale annotations on classification datasets for effective training \cite{NEURIPS2018_4c7a167b}. However, most available datasets in student answer assessments only include score annotations. Providing detailed rationale annotation on existing datasets requires significant domain expert efforts. Furthermore, rational annotations are constrained by the specificity of the information in the dataset, making it difficult to generalise across diverse academic subjects. 

% Recent developments in Large Language Models (LLMs), such as GPT-3 \cite{brown2020language} and ChatGPT\footnote{\url{https://chat.openai.com/chat}} \cite{stiennon2020learning}, have showcased remarkable capabilities not only in generating human-like response but also in exhibiting advanced reasoning and inference abilities. These models have been utilized as "reasoning teachers" to fine-tune smaller language models, harnessing their superior inferential skills while maintaining computational efficiency \cite{ho2022large, lu2022learn}. This approach can potentially overcome the limitations of traditional explanation techniques built on PLM-based text classifiers. LLMs can generate context-aware and nuanced reasoning in an understandable textual format without expensive human annotations. As a result, the LLM-generated rationale for student response assessments offers a great opportunity to efficiently train smaller language models to provide accurate and understandable assessment feedback that is useful for both students and educators, ultimately enhancing the learning experience and assessment explainability.
Recent developments on Large Language Models (LLMs), including ChatGPT \cite{stiennon2020learning}, have demonstrated impressive capabilities in various Natural Language Processing (NLP) applications. For example, these models have exhibited remarkable performance in %Some research indicates that their
arithmetic and common sense reasoning while showing their potential for performing step-by-step %ability can perform 
reasoning %while achieving performance gain 
\cite{Wei2022ChainOT}. Furthermore, \citet{Gilardi2023ChatGPTOC} found that using ChatGPT for data annotation outperforms crowd workers with much lower costs. %All these discoveries shed new light on the tremendous potential of using 
It becomes possible to improve the interpretability of student answer assessment, by harnessing the capabilities of LLMs  %for rationale generation to improve student answer assessment tasks' interpretability 
without relying on expensive human annotation processes. However, LLMs' running costs, non-open-source issues and limited specialization still hinder their applications.

% Introduce our approach
% In this paper, we aim to distill ChatGPT as a reasoning teacher to train a smaller language model to generate meaningful student response assessment rationale via zero/few-shot inference. We developed three prompt templates based on different reasoning levels to compare rationale quality and generate plausible explanations. Moreover, we use semantic confidence from LLMs to study the reliability of human labels and identify corrupted data, and use LLM outputs as noisy label data to fine tune a smaller language model for explainable student answer scoring. %s to do further data augmentation. 
% Extensive experiment results show our method is capable of achieving high student response assessment performance meanwhile provide transparent rationales. This method can not only improve the trustworthiness when applying automated student response assessment systems in large-scale examinations but also shed new light on new strategies to improve transparency in explainable text classification.
% This paper proposed an \textbf{A}utomated \textbf{E}xplainable Student \textbf{R}esponse \textbf{A}ssessment framework, called \textbf{AERA}, to distil ChatGPT as a reasoning teacher to fine-tune a smaller language model to generate rationales and improve the interpretability on student answer assessment tasks. 
This paper introduces the \textbf{AERA} (\textbf{A}utomated \textbf{E}xplainable Student \textbf{R}esponse \textbf{A}ssessment) framework, designed to harness ChatGPT as a reasoning teacher. The aim is to distil a more compact language model, enabling it to produce rationales and enhance interpretability on student answer assessment.
We first designed several prompt templates with different levels of instruction to examine ChatGPT's %reasoning capabilities on answer assessment. We find that zero-shot instructed rationales are prone to a common issue known as the hallucination problem. By incorporating example-based instruction, we significantly reduce the occurrence of hallucinations in the generated rationales. % while instruction with examples can significantly reduce this issue. 
capabilities on student answer assessment and rationale generation. Then, we enhance the quality of rationales with a rationale refinement module. Last, a smaller language model is finetuned on the refined data to perform the answer assessment and rationale generation.
% After identifying the most effective instructing method, we further explore two strategies to enhance the quality of the rationales. 
Since there are no established automatic metrics to evaluate the correctness of rationales without ground truth annotations, we conducted a comprehensive human evaluation, assessing the rationales generated by \textbf{AERA} and comparing them with those generated by ChatGPT. Our experimental results show that, within our designed framework, a smaller language model can surpass ChatGPT in terms of assessment performance %under our designed framework 
while generating more accurate rationales to explain the assessment decision.

% In summary, our contributions are: (1) We proposed various prompt reasoning methods to explore the rationale generation capability of ChatGPT in student response assessment; (2) We adopted semantic confidence interval and data augmentation methods to improve the assessment performance with rationale generation significantly; (3) Comprehensive experiments show that our method is able to generate accurate rationales without requiring human-annotated rationales for model learning. To the best of our knowledge, our proposed method is the first to distill ChatGPT for learning a smaller language models for explainable student response assessment. % without the help of gold rationale label data. 
% We will release our code %and generated rationales 
% on GitHub. % upon paper acceptance.
In summary, our contributions are: (1) We proposed a framework \textbf{AERA}, to distil the rationale generation capability of ChatGPT into a smaller language model; (2) We introduced two strategies for ChatGPT to refine its rationales independently; (3) Through comprehensive experiments and human evaluation, we show that our method is able to generate high-quality rationales without the need of additional annotation for model learning. %To the best of our knowledge, our proposed framework is the first approach to distil ChatGPT for generating rationales for explainable student answer assessment %rationale generation 
% using smaller language models.
To the best of our knowledge, AERA is the pioneering framework that leverages ChatGPT to generate rationales for explainable student answer assessments using more compact language models.

\begin{figure*}[t]
  \centering
  \includegraphics[width=0.95\linewidth]{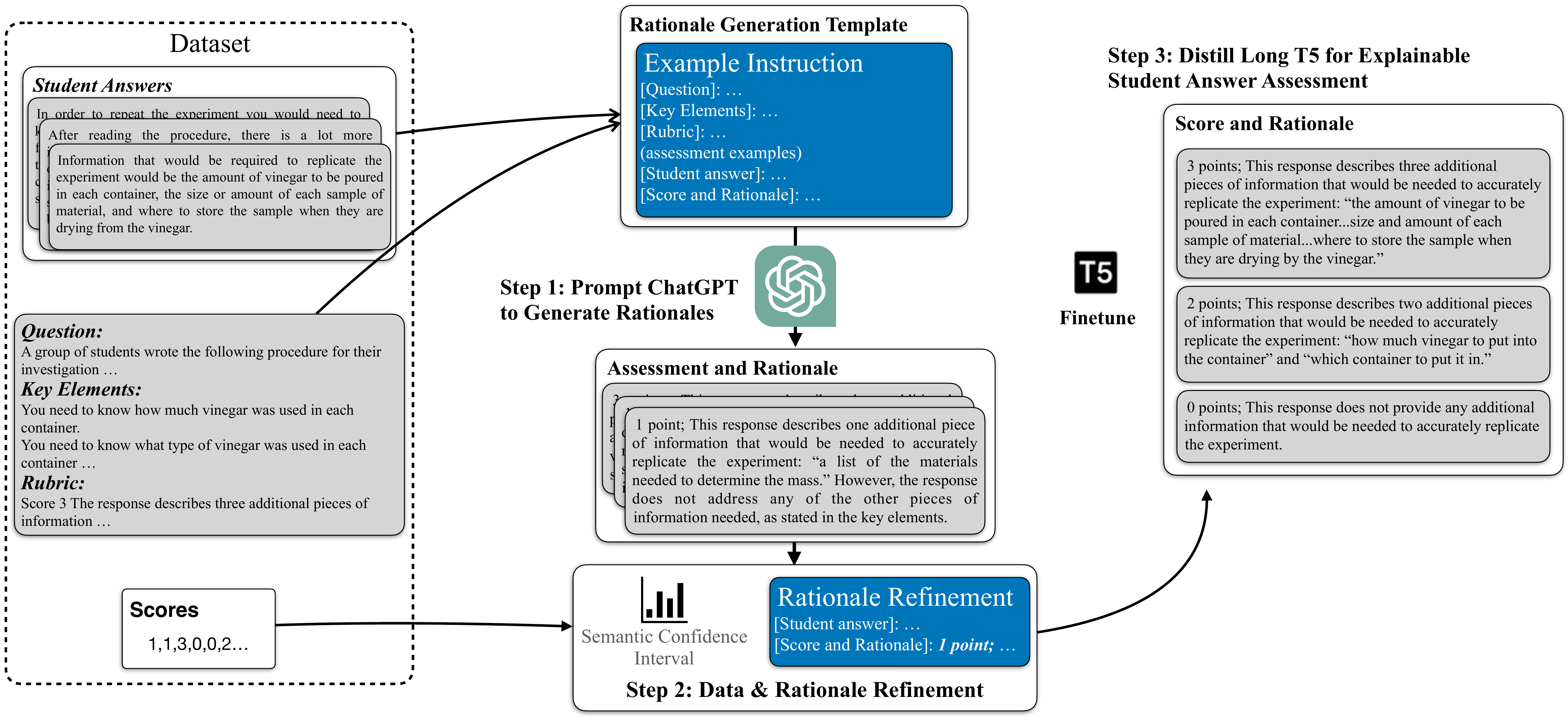}
  \caption{\textbf{AERA} framework contains three steps: (1) Prompting ChatGPT for rationale generation; (2) Applying rationale refinement strategies to improve the quality of the rationales; and (3) Distilling a smaller language model for more efficient rationale generation.}
  \label{fig:aera_framework}
\end{figure*}

\section{Related Work}
% Our research %on using ChatGPT to generate rationale for student response assessment 
% is related to three topics in the literature: automated student answer assessment, % with pre-trained language models, 
% rationale generation, and knowledge distillation. 
% \paragraph{Automated Student Answer Assessment}
% The development of automated student answer assessment systems has been an active area of research in educational technology. Early work in this area focused on employing traditional machine learning algorithms, such as Support Vector Machines %SVM \cite{joachims1998text} 
% and Naive Bayes %\cite{mccallum1998comparison}, 
% for automated essay scoring and short answer grading. With the advent of pre-trained language models, such as BERT \cite{devlin2018bert} and GPT-3 \cite{brown2020language}, researchers have explored their use in various educational applications, including automated essay scoring \cite{filighera-etal-2022-answer}, question-answering \cite{lu2022learn}, and providing feedback on student writing \cite{yannakoudakis-etal-2011-new}. However, most existing approaches suffer from a lack of transparency, as they often produce predictions without providing an explicit rationale behind their decisions. Our work addresses this limitation by leveraging ChatGPT-generated rationales to train a smaller language model for explainable student answer assessment, %generate detailed and understandable rationale for student response assessment, 
% enhancing the trustworthiness and utility of automated assessment systems.
\paragraph{Automated Student Answer Assessment} Also known as automated essay scoring, where most researchers model the problem as a text classification task \cite{Uto2021ARO}. Early approaches \cite{alikaniotis-etal-2016-automatic, dong-etal-2017-attention} built on deep neural networks shed new light on efficient and consistent assessment solutions. Recent advents in PLMs \cite{devlin2018bert,brown2020language} provide better text representations to develop more accurate PLM-based scoring systems \cite{mayfield-black-2020-fine, yang-etal-2020-enhancing}. Nevertheless, limited knowledge of the assessment system's decision-making process raised concerns about its fairness and usefulness. \citet{alikaniotis-etal-2016-automatic, yang-etal-2020-enhancing} tried to improve assessment interpretability via attention mechanisms. %However, their approaches cannot generate natural language explanations, which makes interpretations hard to be used by students or educators directly. 
\citet{filighera-etal-2022-answer} annotated a student feedback dataset for more explainable assessment results with feedback. %Research for interpretable answer assessment systems that generates natural language rationale without human annotation has been less addressed in this area. %Our work addresses limitations on interpretability and annotation cost via leveraging ChatGPT's reasoning ability to train a smaller language model to generate rationales for explainable student answer assessment.

% \paragraph{Rationale Generation}
% Rationale generation in NLP has gained increasing attention in recent years to enhance the interpretability and trustworthiness of NLP models \citep{lei-etal-2016-rationalizing, yu-etal-2019-rethinking}. Rationales are concise, human-readable explanations for model predictions, which can be particularly useful for tasks such as sentiment analysis \citep{lei-etal-2016-rationalizing}, machine translation \citep{wu-etal-2018-phrase}, and document classification \citep{bastings-etal-2019-interpretable}. Methods for generating rationales can be broadly divided into two categories: extraction-based methods, which select a subset of input features as rationales \citep{lei-etal-2016-rationalizing}, and generation-based methods, which synthesize rationales from scratch \citep{liu-etal-2019-towards-explainable, wiegreffe-etal-2021-measuring}. Recent advances in rationale generation have leveraged techniques such as reinforcement learning \citep{yu-etal-2019-rethinking} and pre-trained language models \citep{wiegreffe-etal-2021-measuring} to improve the quality and coherence of the generated explanations. However, challenges still remain in evaluating the quality of generated rationales.
\paragraph{Rationale Generation in Text Classification} Generate rationales for text classifiers have gained increasing attention due to concerns in interpretability \cite{Gurrapu2023RationalizationFE, li2023cue}. Researchers tried to generate rationales on various tasks, including sentiment analysis \cite{antognini-faltings-2021-rationalization}, review classification \cite{liu2019explainable}, and natural language inference \cite{NEURIPS2018_4c7a167b}. Those approaches mainly fall into two categories: extractive rationale generation \cite{lei-etal-2016-rationalizing}, where rationales are extracted from the input features; and abstractive rationale generation \cite{marasovic-etal-2022-shot}, where rationales are paraphrased from existing sentences or newly generated. LLMs showcased the great potential to use their in-context learning ability for abstractive rationale generation \cite{marasovic-etal-2022-shot}, which provides a viable solution for our task. %We focus on generating abstractive rationale with LLMs in a novel domain, student answer assessment.

In our study, we tackle the interpretability challenge in automated student answer assessments by producing abstractive rationales from ChatGPT and distilling a smaller model to perform the same task. %Students or educators can readily utilize these rationales. Furthermore, we fine-tune smaller language models to generate these rationales, ensuring both clarity and competitive assessment results.

% \section{Generating Free-form Rationales for Text Classification via Reasoning Teacher Distillation}
\section{AERA: Automated Explainable Student Response Assessment Framework}
Applications of student answer assessment systems built on PLMs have been hindered by concerns about their interpretability. Existing explanation methods built on classification-based assessment systems struggle to provide natural language explanations for their decision-making process, thus making their application less useful for education purposes. %Moreover, the scarcity of datasets annotated with marking rationales and the high costs associated with human annotation create barriers for developing rationale generation approaches.
Additionally, the limited availability of datasets annotated with grading rationales, coupled with the substantial expenses of human annotation, poses significant obstacles to the advancement of rationale generation approaches.

To address the above challenges, we introduce \textbf{AERA} framework, which leverages the in-context learning capabilities of LLMs to generate rationales and fine-tune smaller language models for explainable student answer scoring. As shown in Figure \ref{fig:aera_framework}, our approach consists of three main steps: (1) We design various prompt templates according to different levels of reasoning difficulties and instruct ChatGPT to assess student answers while providing rationales. (2) It is important to acknowledge that ChatGPT may not be able to assess all student answers accurately. To address this limitation, we introduce a rationale and data refinement module which aims to enhance the quality and usability of the generated rationales. (3) The generated rationales, despite the presence of noise, can be utilized to efficiently fine-tune smaller language models, enabling the generation of plausible rationales for student answer assessment.
% To address these challenges, we introduce a LLM-based, in-context learning framework, \textbf{AERA}: \textbf{A}utomated \textbf{E}xplainable Student \textbf{R}esponse \textbf{A}ssessment, which enables fine-tuning small language models for explainable student answer scoring. %, emphasising explainable text classification in student response assessment. 
% Our approach consists of three main stages: (1) Using ChatGPT for zero-shot or few-shot rationale generation to create an initial set of rationales, (2) Incorporating confidence intervals and predicted results for data selection and further rationale augmentation, and (3) Demonstrating how these generated rationales, despite the presence of noise, can effectively fine-tune small language models for efficient and plausible rationale generation.

% \subsection{Notatisons}
\paragraph{Problem Setup}
A typical student answer assessment dataset includes five components. The question $\mathcal{Q}$\footnote{Some questions contain tabular data. We leverage ChatGPT’s table-understanding capability to create table descriptions from tabular data in \ref{app:tabular_transform}.}; Key elements $\mathcal{K}$ that list the expected key answer elements; Rubrics $\mathcal{R}$, a grading guide used to evaluate the quality of student answers\footnote{See \textsection{\ref{app:prompt_detail}} for detailed questions, key elements and rubrics.}; A collection of student answers $X = \{x_i\}_{i=1}^N$; %|i = 1,2,\cdots, N\}$; 
and a collection of the corresponding scores $Y= \{y_i\}_{i=1}^N$. %Each pair of student answers and scores is an i.i.d. realization of random variables represented as $(x_i, y_i), i = 1, 2,\cdots, N$. 
When preparing the key elements and rubric used for assessment, lead examiners will also provide sample answer assessments during standardisation meetings\footnote{\href{https://www.aqa.org.uk/exams-administration/coursework-controlled-assessment-nea/standardisation}{A standardisation guide from a leading exam service.}}. We denote those sampled student answers, scores and grading rationale as $(x'_j, y'_j, r'_j), j = 1, 2,\cdots, M$. For a given student answer $x_i$, we use $\hat{y}_i$ to denote the predicted score and $\hat{r}_i$ to denote the generated rationale. 

We use the following notations to describe the model generation process: %for a clearer demonstration of rationalization and the model's input and output: 
$X \rightarrow Y$, where the model directly predicts a score given a student answer; $X \rightarrow YR$, where the model predicts a score and generates a rationale given a student answer; $XY \rightarrow R$, where both a student answer and its corresponding score are given to the model to generate a rationale. For the rest of the section, we highlighted \mygreen{examples from sample assessment in green} and \myblue{models' output in blue}.

% \subsection{ChatGPT for Rationale Generation} 
\subsection{Prompting ChatGPT for Rationale Generation} \label{sec:chatgpt_rationale_generation}
Recent advances in ChatGPT showcased its great potential to generate rationales on complex reasoning tasks, such as arithmetic computation. However, student answer assessment is a complex decision-making process in education involving various reasoning phases \cite{bejar_mental_model}. The main challenges for an assessment task include finding the valid key elements stated in the student's answer and deciding a proper score range that applies to the answer. To the best of our knowledge, there is scarce work researching the viable prompting strategy for student answer assessment with ChatGPT. Following the taxonomy from \citet{Karmaker2023TELeRAG}, we propose three prompt templates for rationale generation at different reasoning levels to explore ChatGPT's reasoning capability and identify the prompt leading to more accurate assessment results. %for answer assessment rationalization. 
We begin with the Simple Instruction template, and progressively reduce the level of reasoning difficulty by incorporating more elaborate natural language instructions or patterns extracted from assessment samples into the template.

\paragraph{Simple Instruction} \label{sec:zero-shot-rationale}
We first use a simple $X \rightarrow YR$ prompt instruction that only contains a single question to ask ChatGPT to elaborate the reason for its scoring process:
\begin{lstlisting}[mathescape=true, caption=Simple Instruction Prompt Template]
[Question]: <$\mathcal{Q}$>
[Key Elements]: <$\mathcal{K}$>
[Rubric]: <$\mathcal{R}$>
[Student answer]: <$x_i$>
What score should this Student answer get and why?
(*@\myblue{(free form of $\hat{r}_i$ and $\hat{y}_i$)}@*)
\end{lstlisting}
Given the intricate nature of the student answer assessment task, this prompt presents the highest level of difficulty. ChatGPT needs to plan its assessment cycle, understand the meaning of key elements and the rubric, and appropriately execute the assessment to match the student answer with the key elements and apply the rubric for scoring and rationale generation. 

\paragraph{Complex Instruction} Previous research suggests that more elaborate natural language prompt instruction may improve the reasoning capabilities of LLMs \cite{Karmaker2023TELeRAG, brown2020language}. Therefore, we design a more detailed $X \rightarrow YR$ prompt instruction that clearly outlines the functionality of key elements and the rubric and provides clear guidance on how to apply them in student answer assessment:
\begin{lstlisting}[mathescape=true, caption=Complex Instruction Prompt Template]
[Question]: <$\mathcal{Q}$>
[Key Elements]: <$\mathcal{K}$>
[Rubric]: <$\mathcal{R}$>
[Student answer]: <$x_i$>
Carefully read the [Question], [Key Elements], and [Rubric], then compare [Student answer] with the [Key Elements], and apply the [Rubric] to derive the student score. Please be certain to spell out your reasoning so anyone can verify them. Spell out the [Key Elements] that the [Student answer] matches, and also spell out which rule in the [Rubric] is applied.
(*@\myblue{(free form of $\hat{r}_i$ and $\hat{y}_i$)}@*)
\end{lstlisting}
Compared with the \textit{Simple Instruction} template, the \textit{Complex Instruction} template offers additional guidance in the assessment process, thereby reducing the level of difficulty.   

\paragraph{Example Instruction} Although ChatGPT has demonstrated impressive reasoning capabilities in understanding natural language instructions, it faces some limitations when employing zero-shot based templates such as the aforementioned \emph{Simple} and \emph{Complex Instructions}.  Specifically, it tends to generate free-form rationales that require additional annotations for score and rationale extraction. In addition, it also suffers from hallucination problems. Previous research \cite{brown2020language, marasovic-etal-2022-shot,openai2023gpt4, Karmaker2023TELeRAG} has shown the benefits of few-shot based templates, as they allow output formatting through examples, eliminating the need for annotations for score extraction. Furthermore, leveraging the patterns from demonstration examples, LLMs can achieve better performance. To this end, we proposed a $X \rightarrow YR$ example instruction prompt, utilizing the sample answer assessments obtained from standardisation as demonstration examples, for generating properly formatted rationales:
\begin{lstlisting}[mathescape=true, caption=Example Instruction Prompt Template]
[Question]: <$\mathcal{Q}$>
[Key Elements]: <$\mathcal{K}$>
[Rubric]: <$\mathcal{R}$>
(*@\mygreen{[Student answer]: <$x'_1$>}@*)
(*@\mygreen{[score and Rationale]: <$y'_1$>; <$r'_1$>}@*)
(*@\mygreen{...(assessment examples)}@*)
(*@\mygreen{[Student answer]: <$x'_M$>}@*)
(*@\mygreen{[score and Rationale]: <$y'_M$>; <$r'_M$>}@*)
[Student answer]: <$x_i$>
[score and Rationale]:(*@ \myblue{<$\hat{y}_i$>; <$\hat{r}_i$>}@*)
\end{lstlisting}

% \paragraph{Tabular Data Transformation}
% \begin{figure}[t]
%   \centering
%   \includegraphics[width=\linewidth]{table_transformation.pdf}
%   \caption{Demonstration of using ChatGPT for tabular data and table description transformation.}
%   \label{fig:table_transform}
% \end{figure}
% Some questions in our dataset contain tabular data, which poses a challenge for smaller language models in terms of inputting and understanding structured data. To address this issue, as shown in Figure \ref{fig:table_transform}, we leverage ChatGPT's table-understanding capability to create table descriptions from the tabular data and verify the description correctness by having ChatGPT generate a table based on the description\footnote{Obtained with ChatGPT version 13th Feb 2023.}. Notably, we found that all the tabular data in our dataset could be accurately reconstructed based on the description generated by ChatGPT. Consequently, we replaced all the tabular data from the question part of our prompts with the corresponding generated descriptions.

\subsection{Data \& Rationale Refinement} \label{sec:data_refinement}

Given the lack of established approach to evaluate the correctness of generated rationales without gold annotation, we follow a previous study \cite{ho2022large} by assuming the rationale supports the score if the LLM-predicted answer score is correct. However, it is important to note that ChatGPT cannot guarantee the correctness of all the assessed scores on the whole dataset. Incorrect predictions can arise from two scenarios: (1) The dataset contains wrongly labelled score; or (2) ChatGPT's predictions are wrong. To address these situations, we introduce refinement strategies to improve the rationale generation's success rate.

\paragraph{Fixing Wrongly Labelled Data} ChatGPT, being a non-deterministic language model, can generate varying outputs with each iteration. We utilise the semantic confidence interval for LLMs outlined by \citet{kuhn2023semantic} to calculate the uncertainty of scores associated with the generated rationales. Based on our observation, generated rationales $\hat{r}_i$ that correspond to the same assessed score $\hat{y}_i$ are semantically similar. Therefore, the predictive probability of each assessed score $\hat{y}_i$ can be represented as: $p(\hat{y}_i\mid x_i)=\sum_{\hat{y}_i\in S} p(\hat{y}_i\mid x_i)$; where $S$ is the set of all occurrences of semantically similar rationales shares the same predicted score. 

Through our experiments, we demonstrate that gold annotations might be wrong for highly confident incorrect assessments made by ChatGPT, when the score difference exceeds one. This approach helps to identify corrupted input data and human labelling errors, ultimately reducing data uncertainty and improving overall data quality.

% \subsubsection{Further Rationale Refinement} \label{sec:rationale_refinement}
\paragraph{Prompt for Rationale Refinement} Since the $X\rightarrow YR$ prompt cannot guarantee the correctness of the score, we introduce a $XY\rightarrow R$ rationale refinement template. This template is based on the \emph{Example Instruction} prompt template and incorporates a given score as input, LLM can use the score as prior knowledge to locate a proper distribution that generates more accurate rationales:
\begin{lstlisting}[mathescape=true, caption=Rationale Refinement Prompt Template]
[Question]: <$\mathcal{Q}$>
[Key Elements]: <$\mathcal{K}$>
[Rubric]: <$\mathcal{R}$>
(*@\mygreen{[Student answer]: <$x'_1$>}@*)
(*@\mygreen{[score and Rationale]: <$y'_1$>; <$r'_1$>}@*)
(*@\mygreen{...(assessment examples)}@*)
(*@\mygreen{[Student answer]: <$x'_E$>}@*)
(*@\mygreen{[score and Rationale]: <$y'_E$>; <$r'_E$>}@*)
[Student answer]: <$x_i$>
[Score and Rationale]: <$y_i$>;(*@ \myblue{<$\hat{r}_i$>}@*)
\end{lstlisting}
\subsection{Distilling Student Model for Efficient Rationale Generation}
Although LLMs have exhibited impressive in-context learning and reasoning capabilities, huge parameter size, non-open source issues, and enormous running costs \cite{chatgpt_cost, Li2023OverPromptEC} make them hard to be developed and trained locally. Besides, uncontrollable, occasionally unexpected outputs (e.g. hallucination) render LLMs less practical for real-world student answer assessment. Consequently, we propose using ChatGPT-generated rationales to fine-tune a smaller language model for efficient rationale generation. Unlike previous literature that has focused on knowledge distillation in arithmetic chain-of-thought \citep{ho2022large, pmlr-v202-fu23d,magister-etal-2023-teaching}, student answer assessment is a much more complex reasoning task based on the input source (e.g. the scope of key elements and the definition of the rubric). %On the other hand, arithmetic is more of a common-sense knowledge and appears more frequently in most of the pre-training datasets.

\begin{table*}[ht]
\centering
\resizebox{\linewidth}{!}{
\begin{tabular}{@{}lccc|ccc|ccc|ccc|ccc@{}}
\toprule
\textbf{Dataset} (Subject)   & \multicolumn{3}{c}{\textbf{\#1} (Science)} & \multicolumn{3}{c}{\textbf{\#2} (Science)} & \multicolumn{3}{c}{\textbf{\#5} (Biology)} & \multicolumn{3}{c}{\textbf{\#6} (Biology)} & \multicolumn{3}{c}{\textbf{Overall}} \\
\textbf{Method/Model}   &  Acc & F1 & QWK &  Acc & F1 & QWK &  Acc & F1 & QWK &  Acc & F1 & QWK &  Acc & F1 & QWK  \\\midrule 
 \multicolumn{16}{c}{$X\rightarrow Y$ Fine-tuned Text Classification} \\\midrule
BERT & 66.79 & 67.54 & 79.17 & 54.23 & 51.53 & 68.53 & 84.28 & 45.82 & 72.87 & 88.43 & 54.76 & 80.30 & 73.43 & 54.91 & 75.22 \\
% \multicolumn{16}{c}{$X\rightarrow Y$ Fine-tuned Generative Text Classification} \\\midrule
% Long T5 (Generative)&68.58 & 69.53 & 80.89 & 59.31 & 58.92 & \textbf{75.61} & 85.34 & 56.38 & 76.29 & 88.37 &58.38 &81.68 & 75.40 & 60.80 & 78.62\\
% Long T5-all (Generative)&69.96 &70.84 & \textbf{82.09} & 56.57 &53.12 &57.83 & 85.62 & 58.94 & 79.53\\\midrule
Longformer&74.15 & 74.81 & \textbf{83.75} & 62.75 & 63.21 & \textbf{78.79} & 83.67 & 58.02 & 80.63 & 88.09 & 59.44 & \textbf{83.25} & 77.17 & 63.87 & \textbf{81.61}\\
Longformer-all & 72.59 & 73.61 & 83.05 & 59.08 & 59.52 & 76.76 & 86.23 & 61.50 & \textbf{82.17} & 87.59 & 55.82 & 82.56 & 76.37 & 62.61 & 81.14 \\\midrule 
 \multicolumn{16}{c}{$X \rightarrow YR$ ChatGPT Prompting} \\ \midrule
Simple Instruction& 49.19 & 46.19 & 58.69 & 46.86 & 43.49 & 56.11 & 53.01 & 41.48 & 42.76 & 43.91 & 29.61 & 41.14 & 48.24 & 40.19 & 49.68 \\
Complex Instruction& 55.30 & 55.28 & \underline{65.38} & 38.82 & 38.33 & 45.06 & 71.24 & 41.26 & 52.94 & 70.78 & 52.06 & \underline{64.73} & 59.04 & 46.73 & 57.03\\
Example Instruction& 55.66 & 53.75 & 61.40 & 49.06 & 48.12 & \underline{63.20} & 68.06 & 54.02 & \underline{68.17} & 68.45 & 50.39 & 64.66 & 60.31 & 51.57 & \underline{64.36}\\\midrule
 \multicolumn{16}{c}{$X \rightarrow YR$ Fine-tuned Long T5 Rationalization} \\ \midrule
\textbf{AERA}(\textbf{Ours}) & 63.26 & 62.90 & \textbf{75.06} & 43.27 & 42.35 & \textbf{54.15}& 83.78 & 53.29 & \textbf{76.44} & 89.37 & 60.38 & \textbf{80.81} & 69.92 & 54.73 & \textbf{71.62}\\
w/o Fixing Wrong Labels & 52.42 & 50.24 & 60.66 & 40.45 & 35.29 & 44.26& 66.78& 50.76& 63.65& 68.00& 40.94& 62.54&55.91&44.31&57.78\\
w/o Rationale Refinement & 52.06 & 49.46 & 58.95 & 40.85 & 39.09 & 49.80 & 61.65& 46.47& 60.18& 66.94& 41.18& 62.05& 55.38& 44.05&57.75\\
Correct Score Only & 56.79 & 55.95 & 69.96 & 35.60 & 24.02 & 23.94 & 78.76 & 48.79 & 71.83 & 84.19 & 54.93 & 79.17 & 63.84 & 45.92 & 61.23\\
\bottomrule
\end{tabular}}
\caption{Comparison of performance across classification baselines and rationale generation approaches. The highest QWK has been highlighted in \textbf{Bold} for fine-tuned models and \underline{underlined} for LLM inference results.}% \dag indicates trained with train sets filtered by incorrect ChatGPT-predicted answer scores.}
% \ddag stands for trained with a train set with additional refined rationales and fixed incorrect labels.}
\label{tab:model_comparison}
\end{table*}

We utilise the rationales generated by ChatGPT, as described in \textsection{\ref{sec:chatgpt_rationale_generation}}, with their quality improved by fixing wrongly labelled data and further refinement outlined in \textsection{\ref{sec:data_refinement}}, as training data for task-specific knowledge distillation. We adopt Long T5 \cite{guo-etal-2022-longt5} as our base model as T5 is one of the popular open-source PLM that has been pre-trained with many supervised tasks, including both classification and generation. % and rationalization on student answer assessment requires both classification and text generation. 
Besides, prompt for student answer assessment is relatively long, Long T5 is capable of taking longer input than commonly used base models while maintaining little performance drop.
Our fine-tuning process takes Question, Key Elements, Rubric, and student answer as input to predict the score and generate rationale, $X \rightarrow YR$. %We adopt sacreBLEU to evaluate the rationale's semantic similarity on the validation set and choose the best checkpoint. 
Prompt template used for fine-tuning is as follows:
\begin{lstlisting}[mathescape=true, caption=Prompt Template for Fine Tuning]
[Question]: <$\mathcal{Q}$>
[Key Elements]: <$\mathcal{K}$>
[Rubric]: <$\mathcal{R}$>
[Student answer]: <$x_i$>
[Score and Rationale]:(*@ \myblue{<$\hat{y}_i$>; <$\hat{r}_i$>}@*)
\end{lstlisting}

\section{Experiments}
\subsection{Experimental Setup}
\paragraph{Dataset} We employ the Hewlett Foundation: Short Answer Scoring (ASAP-SAS) dataset\footnote{\url{https://kaggle.com/competitions/asap-sas}}. This dataset encompasses over 23,000 short answer responses from students in grades 7 to 10, including ten questions spanning subjects such as Science, Biology, English, and Art. We only use four subsets focusing on Science and Biology questions.
\paragraph{Baselines}
We compare our method with three classification setups: \textbf{BERT}: Bert-base-uncased model fine-tuned with student answers as inputs and scores as output \cite{mayfield-black-2020-fine}; \textbf{Longformer}: Longformer-base-4096 fine-tuned with student answers as input and scores as output; and \textbf{Longformer-all}: Longformer-base-4096 fine-tuned with the concatenation of additional information (question, key elements, rubric) and student answers as input and scores as output. %\textbf{Long T5}: Long-t5-tglobal-large fine-tuned with student answers as input and scores as output; \textbf{Long T5-all}: Long-t5-tglobal-large fine-tuned with the concatenation of additional information (question, key elements, rubric) and student answers as input and scores as output. 
\paragraph{Evaluation Metric}
We adopt the Accuracy (Acc) and macro f1 score (F1) and Quadratic Weighted Kappa (QWK) to evaluate the classification performance. We use sacreBLEU \cite{post-2018-call} to measure the rationales' semantic similarity on the validation set and select the best checkpoint.

\noindent We provide detailed dataset description, QWK implementation and hyper-parameters setup in \textsection{\ref{app:experimental_setup}}.

\subsection{Overall Comparison}
Table \ref{tab:model_comparison} displays the performance of student answer assessment across three task scenarios: fine-tuned text classification, ChatGPT prompting, and fine-tuned Long T5 for rationale generation.

For text classification baselines, when comparing BERT and Longformer, we observe that using a model that accommodates longer input text length can improve performance when trained solely on student answers. However, we do not see an improvement in overall performance when incorporating additional information, such as question, key answer elements and rubric, into the input, which suggests that the text classifier  
%Instead, three out of four datasets experience a performance decline, suggesting that these resources may be treated as noise and not utilized in the student assessment process. Therefore, text classifier-based student response assessment 
may make predictions based on spurious features rather than checking student answer against the key answer elements and applying the supplied rubric. %not be reliable since it didn't use appropriate resources in assessment, thus making it 
Hence, even though text classifiers may exhibit relatively high-performance scores, there remains a concern about the trustworthiness of their outputs.
%less trustworthy.

%For results generated from ChatGPT, we observe that using a more elaborated instruction gives better results compared to a simple instruction. We observe performance improvement in two of four datasets by providing some demonstration examples and even without using elaborate instructions, leading to the highest overall performance scores and the lowest variance across all four datasets. This shows the superiority of the few-shot (Example Instruction) compared to the zero-shot (Simple \& Complex Instruction) settings. We can conclude that, like human examiners, ChatGPT provided with assessment samples can better apply the rubric and key elements, making it the most appropriate prompt strategy for rationale generation on automated student answer assessment.
For assessment and rationale generated from ChatGPT, we observe that the prompting under the few-shot setting (\emph{Example Instruction}) is superior to the zero-shot settings (\emph{Simple} \& \emph{Complex Instruction}), which achieved the highest overall performance with lower variances across four datasets.

Once we identified the viable prompt strategy for rationale generation, we fine-tuned Long T5 on the generated rationale for explainable student answer assessment. Our \textbf{AERA} framework obtained the highest overall performance compared with other rationale generation methods. Although the overall performance does not match that of text classifiers, given the intricate nature of the text generation task, noteworthy performance gains are observed on datasets \#5 and \#6, surpassing those achieved by the BERT classifier. This shows the benefit of enhancing the transparency of automated student answers assessment by generating rationales.

We conducted ablation studies to examine the effectiveness of each component in the Data \& Rationale Refinement module in our framework. We find that if we only keep a subset of rationales with correctly predicted scores\footnote{See \textsection{\ref{app:filtered}} for detailed data statistic and other comparisons.} (\emph{Correct Score Only}), the performance on datasets \#1, \#5 and \#6 surpasses those achieved when incorporating any of the additional refinement strategies. Although these results show the strong performance brought by rationales with correctly predicted scores, this method may not be universally applicable when the amount of data is limited, as seen in dataset \#2. After incorporating the two strategies separately, namely Fixing Wrong Labels \& Rationale Refinement, to compose an updated dataset, we observed a significant performance improvement on dataset \#2 due to the availability of more data. However, we see a performance drop on \#1, \#5 and \#6 when compared with \emph{Correct Score Only}, indicating the presence of wrongly labelled data or incorrectly predicted rationales can adversely impact the overall performance. In sum, both the data \& rationale refinement components are essential within our framework to prevent data scarcity and effectively reduce noisy data.

\subsection{Human Evaluation}
We carried out two distinct human evaluations for rationales generated by both AERA and ChatGPT\footnote{For a comprehensive evaluation setup, data statistics, IAA scores, and various breakdown evaluation results, refer to \textsection{\ref{app:human_eval}}}.
\begin{figure}[ht]
\centering
\includegraphics[width=\linewidth]{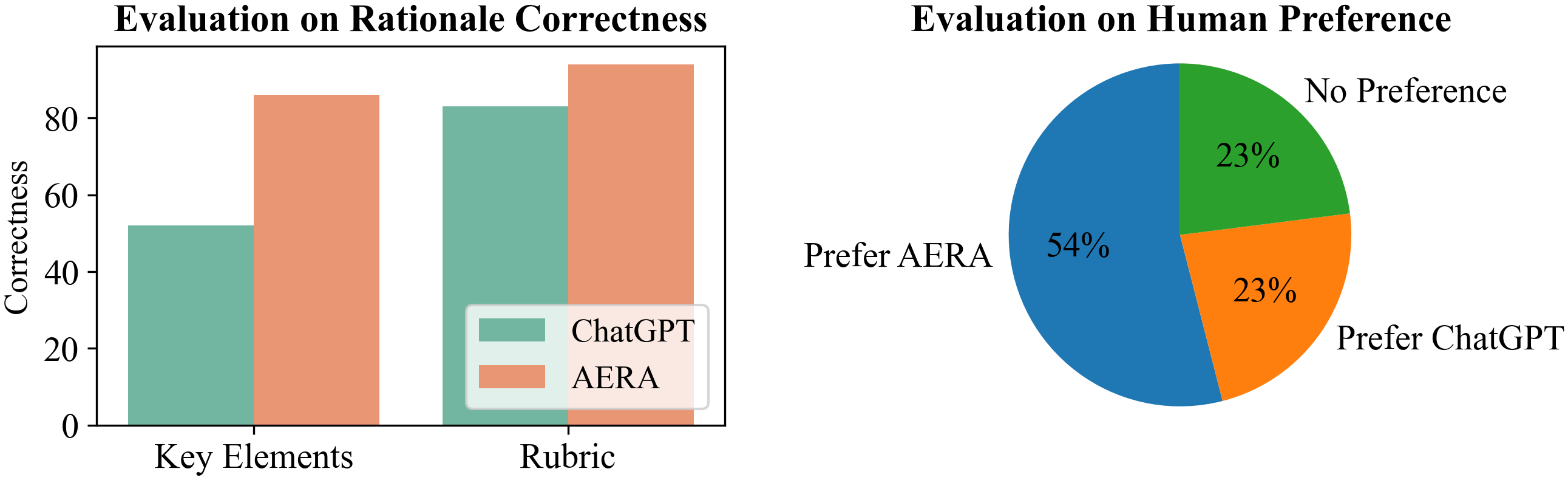}
\caption{Visualization of Human Evaluation Results}
\label{fig:stage1_human_evaluation}
\end{figure}

\paragraph{Rationale Correctness}
The initial evaluation centred on the accuracy of rationales. Annotators evaluated the rationales based on two primary criteria: (1) Correctness of matched key elements: Evaluating whether the rationale correctly identifies key elements mentioned by the student's answer. (2) Faithfulness of rubric application: Reviewing if the used rubric corresponds appropriately with the score assigned to the student's answer and the elements identified in the rationale.
\paragraph{Preference on Rationale}
The subsequent evaluation was tailored towards annotators' preferences concerning the rationales. Annotators were shown rationales generated by AERA and ChatGPT in a randomized order. Their task was to choose the rationale they deemed superior, basing their decision on factors such as accuracy and informativeness. The chosen rationale was gauged on its ability to aptly convey pertinent details and its efficacy in substantiating the student answer assessment.

\paragraph{Overall Analysis}
The left segment of Figure \ref{fig:stage1_human_evaluation} indicates that AERA-generated rationales considerably surpass ChatGPT-generated ones across both evaluation criteria. Given the inherent challenge for language models to pinpoint key elements resonating with student answers, it's noticeable that scores tend to be lower for the correctness of matched key elements compared to the rubric application's faithfulness for both models.

The right segment of Figure \ref{fig:stage1_human_evaluation} underscores a marked inclination among annotators towards AERA-generated rationales over ChatGPT's. Despite LLMs sometimes offering more expansive explanations due to their in-context learning prowess, they frequently underperform in accurately gauging student answers relative to our refined model, leading to a diminished preference rating.

% illustrates the evaluation results from both stages. As shown in the left sub-figure, AERA exhibits less accurate matching on Key Elements compared to ChatGPT, but demonstrates a higher level of faithfulness in the application of Rubric. The mismatch in key elements often arises from AERA's tendency to select different key elements in explanation rather than the corresponding ones. As shown in the right sub-figure, annotators displays a greater preference for ChatGPT-generated rationales compared to those generated by AERA. This is mostly because LLMs possess the ability to generate more expansive explanations due to their in-context learning capability and rich knowledge encoded through pre-training. 
% Overall, the smaller language model distilled with the AERA framework is able to surpass ChatGPT's student answer assessment ability while generating more accurate rationales despite being a few orders of magnitude smaller than ChatGPT.
In summary, the compact language model distilled using the AERA framework not only outperforms ChatGPT in student answer assessment but also produces more precise rationales, despite its significantly smaller size compared to ChatGPT.

\subsection{Case Studies on Refinement Strategies}
\begin{table}[ht!]
  \centering
  \resizebox{\linewidth}{!}{
  \small
  \begin{tabular}{ll}
    \toprule
    \multicolumn{2}{p{0.99\linewidth}}{\textbf{\#1}: In order to replicate this procedures, you would need to} \\% know} \\
    \textbf{Original Label}: 3 & \textbf{High Confident Prediction}: 0 \\
    \multicolumn{2}{p{0.99\linewidth}}{\textbf{Rationale}: This response is incomplete and does not provide any relevant information.} \\
    \midrule
    \multicolumn{2}{p{0.99\linewidth}}{\textbf{\#2}: In conclusion trial any} \\
    \textbf{Original Label}: 2 & \textbf{High Confident Prediction}: 0 \\
    \multicolumn{2}{p{0.99\linewidth}}{\textbf{Rationale}: The student’s response is incomplete and does not provide a valid conclusion or any ways to improve the experimental design and/or the validity of the results.} \\
    % \midrule
    % \multicolumn{2}{p{0.97\linewidth}}{\textbf{\#2}: Plastic} \\
    % \textbf{Original Label}: 3 & \textbf{High Confident Prediction}: 0 \\
    % \multicolumn{2}{p{0.97\linewidth}}{\textbf{Rationale}: The response is incomplete and does not address the question prompt.} \\
    \midrule
    \multicolumn{2}{p{0.99\linewidth}}{\textbf{\#1}: In order to replicate this experiment, you would need to know:  1. \myorange{how they got the mass of the four different samples} 2. A list of constants  3. You would have to \myorange{know how much of a sample you would place into the container of vinegar and if it's the same for all four materials}.} \\
    \textbf{Original Label}: 0 & \textbf{High Confident Prediction}: 2 \\
    \multicolumn{2}{p{0.99\linewidth}}{\textbf{Rationale}: This response describes two additional pieces of information that would be needed to accurately replicate the experiment: “how they got the mass of the four different samples” and “you would have to know how much of a sample you would place into the container of vinegar and if it's the same for all four materials.” The response also mentions a list of constants, but it is not clear what specific additional information is being requested.} \\
    \bottomrule
  \end{tabular}}
  \caption{Examples of Incorrectly Labeled Data.}
  \label{tab:confidence_aleatoric}
\end{table}

\begin{table*}[!ht]
  \centering
  \resizebox{0.99\linewidth}{!}{
  \small
  \begin{tabular}{ll}
    \toprule
    \multicolumn{2}{p{0.97\linewidth}}{\textbf{\#1}: \textit{The students data needed to include how much of the solution was poored. They also needed to find out if the rocks weighed the same when they started.}} \\
    \multicolumn{2}{p{0.97\linewidth}}{\textbf{Original}: 1 point; This response describes one piece of relevant information that would be needed to accurately replicate the experiment: “how much of the solution was poured.”}\\
    \multicolumn{2}{p{0.97\linewidth}}{\textbf{Refined}: \textit{0 points;} This response describes little or no accurate or relevant information from the acid rain investigation.} \\
    \midrule
    \multicolumn{2}{p{0.97\linewidth}}{\textbf{\#5}: \textit{There four major steps of protien synthesis. The first is the mRNA going to the rRNA. Then the rRNA takes the information from the mRNA to the tRNA. Next the tRNA will take the information it has and translate it. Lastly, the tRNA will take the information and make a protein.}} \\
    \multicolumn{2}{p{0.97\linewidth}}{\textbf{Original}: 2 points; The student answer matches two key elements, “...mRNA going to the rRNA...” and “...tRNA will take the information and make a protein...”. However, the other two steps are not described accurately or comprehensibly.}\\
    \multicolumn{2}{p{0.97\linewidth}}{\textbf{Refined}: \textit{1 point;} The student answer matches only one key element, “...mRNA going to the rRNA...”} \\
    \midrule
    \multicolumn{2}{p{0.97\linewidth}}{\textbf{\#6}: \textit{1.-Osmosis is how water gets diffused thru the membrane. 2.-Active Transport is where an enzyme opens the cell membrane for an object to come in, and extra energy is needed to assist the object into the cell.3.-Passive Transport is where an enzyme opens the cell, but the object doesn't need the extra energy to come in.}} \\
    \multicolumn{2}{p{0.97\linewidth}}{\textbf{Original}: 2 points; This student answer matches two key elements, “Osmosis... how water gets diffused thru the membrane” and “Passive Transport... enzyme opens the cell, but the object doesn't need the extra energy to come in”. However, the explanation for “Active Transport...” is not entirely accurate, as it mentions an enzyme opening the membrane, which is not correct. Also, the explanation doesn't match the entire description of active transport.}\\
    \multicolumn{2}{p{0.97\linewidth}}{\textbf{Refined}: \textit{3 points;} This student answer matches three key elements, “Osmosis... how water gets diffused”, “Active transport... enzyme opens the cell membrane for an object to come in, and extra energy is needed” and “Passive transport... enzyme opens the cell, but the object doesn't need the extra energy to come in”.} \\
    \bottomrule
  \end{tabular}}
  \caption{Improved Rationale Examples Using the Rationale Refinement Strategy.}
  \label{tab:ration_refine_study}
\end{table*}

\paragraph{Identification of Incorrectly Labeled Data}
% We can derive a confidence score for each instance in the training set using the semantic confidence interval method outlined in \textsection{\ref{sec:confidence_interval}}. ChatGPT may make errors similar to human assessors, who may count fewer or more key elements that correspond to the student response, leading to a one-score difference from the gold label for most incorrect predictions in the training set. Instead, we focus on analyzing the 100\% confident incorrect predictions with a score difference greater than one, as illustrated in Table \ref{tab:confidence_aleatoric}.
As shown in Table \ref{tab:confidence_aleatoric}, we discover that highly confident incorrect predictions by ChatGPT may actually be correct using the method outlined in \textsection{\ref{sec:data_refinement}}, suggesting that the data may be noisy or mislabelled. For example, in the first two cases, student answers are incomplete, possibly due to data corruption. The discrepancy between the human labels and the actual answers highlights the clear mismatch in the original dataset. %yet the original score assigns a relatively high score to answers. 
Besides, we also identify instances that may have been annotated with incorrect lower scores. For instance, the last example in the table clearly covers two key elements based on the rubric (\myorange{highlighted in orange}), but the original score given is 0 point. Such mislabeled data could be difficult to detect without manual examination in a text classification setup. %, as we might lack the resources needed for identification. 
The above discoveries from the dataset, which have not been highlighted in previous research, serve as a validation of our concern regarding the presence of inconsistent marking standards in large-scale student answer assessments. Our approach provides a feasible solution to automatically identifying label inconsistency or data corruptions in a human-annotated dataset.
%. This inconsistency can lead to incorrect student feedback, 
%emphasizing the need for an explainable automated student assessment system.

\paragraph{Rationale Refinement}  As shown in Table \ref{tab:ration_refine_study}, we demonstrate that by providing the \textit{correct assessment points} in the prompt, ChatGPT is able to improve its generated rationale better aligned with the score provided. For example, in the first case, an incorrect key element was initially identified. However, after the correct score is provided to ChatGPT, the model is able to correctly trace back the applicable rubric and thus decides that no key elements were mentioned in the text. We discovered that the original incorrect identification might have been influenced by the presence of "\emph{Other acceptable responses}" stated in the key elements. Determining which part of the response falls into the "\emph{acceptable}" category can be challenging for ChatGPT. The other two examples demonstrated common mistakes in human annotations that occurred in the dataset. In these two cases, ChatGPT might have misinterpreted some student descriptions, but the refinement step is able to rectify the mismatches in key elements. However, this strategy cannot be applied if the data contains wrongly labelled instances, as ChatGPT will be forced to generate rationales that may not make sense. Given the above observations, we urge the need for future development of student answer assessment datasets to provide enough examples for key elements. This could help mitigate ambiguous definitions and provide clearer guidelines for key elements, thereby reducing confusion and improving the consistency of the student answer assessment process.\footnote{We provide further experimental details and comprehensive ablation studies in \textsection{\ref{app:experiments}}.}

% \noindent We provide further experimental details, analysis on ChatGPT-generated rationales, hallucination analysis and comparison of generated rationales in \textsection{\ref{app:experiments}}.

\section{Conclusion}
% In this paper, we investigate the application of ChatGPT in student response assessment, demonstrating the potential of ChatGPT-generated rationales for providing explainable and transparent evaluations. We propose a rationale generation framework AERA and examine three prompt strategies with varying levels of reasoning difficulty, revealing that few-shot based prompting methods are best suited for generating plausible and accurate rationales. The utilization of ChatGPT's semantic confidence interval reveals that LLM results may be more precise than human annotations, underscoring the necessity for automated response assessment models. Further experimental outcomes indicate that fine-tuned student models can offer more accurate student response assessments while generating meaningful rationales, which can be applied to provide responsive student feedback. This paper presents a paradigm for harnessing LLMs' capabilities in student response assessment with minimal fine-grained data to decrease potential human assessor costs and enhance the reliability and transparency of large-scale examinations. 
In this paper, we proposed a framework called \textbf{AERA}, which leverages the in-context learning and reasoning capabilities of ChatGPT for rationale generation in student answer assessment. Our experimental results suggest that although ChatGPT is able to generate free-form rationales with natural language instructions, the example instructed prompt strategy achieves the best performance. We further demonstrate \textbf{AERA} can effectively distil a smaller language model for efficient rationalization on automated student answer assessment tasks, without the need for additional human annotation on rationales. Extensive experiments and human evaluation results have validated the efficacy of the refinement module, and our distilled language model can outperform the teacher model in grading while providing reliable rationales. Our approach presents a cost-effective and efficient method for explainable automated student answer assessment.

% Future work will focus on developing automated metrics for rationale evaluation without annotated data, exploring generalizability, and reducing dependency on ChatGPT by fine-tuning open-sourced LLMs.
% \newpage

\section*{Limitations}
This study has several limitations. First, there may be variations in the designs of prompt templates among individuals, and the manual prompt performance can differ across different datasets. Moreover, due to the extensive search space involved in generating automated prompt text, the auto prompt approach cannot be adequately tested with our current computational resources. Second, although appropriate training has been provided for the annotators, the lack of background in exam assessment among the human evaluation annotators may have some impact on the quality of the evaluations. Lastly, we identified a trade-off between interpretability and assessment performance. Given the variations in base models and structures, bridging this gap remains challenging at present.
% EMNLP 2023 requires all submissions to have a section titled ``Limitations'', for discussing the limitations of the paper as a complement to the discussion of strengths in the main text. This section should occur after the conclusion, but before the references. It will not count towards the page limit.  

% The discussion of limitations is mandatory. Papers without a limitation section will be desk-rejected without review.
% ARR-reviewed papers that did not include ``Limitations'' section in their prior submission, should submit a PDF with such a section together with their EMNLP 2023 submission.

% While we are open to different types of limitations, just mentioning that a set of results have been shown for English only probably does not reflect what we expect. 
% Mentioning that the method works mostly for languages with limited morphology, like English, is a much better alternative.
% In addition, limitations such as low scalability to long text, the requirement of large GPU resources, or other things that inspire crucial further investigation are welcome.

\section*{Ethics Statement}
The dataset utilized in this study is an open-source collection of anonymous student responses, and does not contain any sensitive or identifiable information. Although we have not identified any harmful outputs from ChatGPT in our study, it is worth noting that previous research has observed instances where ChatGPT produced unexpected results. We encourage other researchers to utilize this framework to scrutinize the output generated from specific prompts in ChatGPT that may have the potential to generate harmful information. %While our study includes comprehensive experiments and evaluations by human reviewers, the implementation of an automated system for assessing student responses necessitates further, more detailed evaluation. Particular attention must be paid to the balance of data used, as imbalances could give rise to misleading patterns and potentially unjust assessment results in real-world applications.

% Scientific work published at EMNLP 2023 must comply with the \href{https://www.aclweb.org/portal/content/acl-code-ethics}{ACL Ethics Policy}. We encourage all authors to include an explicit ethics statement on the broader impact of the work, or other ethical considerations after the conclusion but before the references. The ethics statement will not count toward the page limit (8 pages for long, 4 pages for short papers).

\section*{Acknowledgements}

This work was supported in part by the UK Engineering and Physical Sciences Research Council (grant no. EP/T017112/2, EP/V048597/1, EP/X019063/1). JL is funded by a PhD scholarship provided by AQA. YH is supported by a Turing AI Fellowship funded by the UK Research and Innovation (grant no. EP/V020579/2).  

\bibliography{custom_formatted}
\bibliographystyle{acl_natbib}
% \begin{figure}[t]
%   \centering
%   \includegraphics[width=\linewidth]{model_plot.png}
%   \caption{Comparison between models on average QWK performance based on model parameter size.}
%   \label{fig:model_size}
% \end{figure}

\clearpage

\appendix
\setcounter{table}{0}
\renewcommand{\thetable}{A\arabic{table}}
\setcounter{figure}{0}
\renewcommand{\thefigure}{A\arabic{figure}}

\section{Further Framework Details}

\subsection{Tabular Data Transformation} \label{app:tabular_transform}

\begin{figure}[h]
  \centering
  \includegraphics[width=\linewidth]{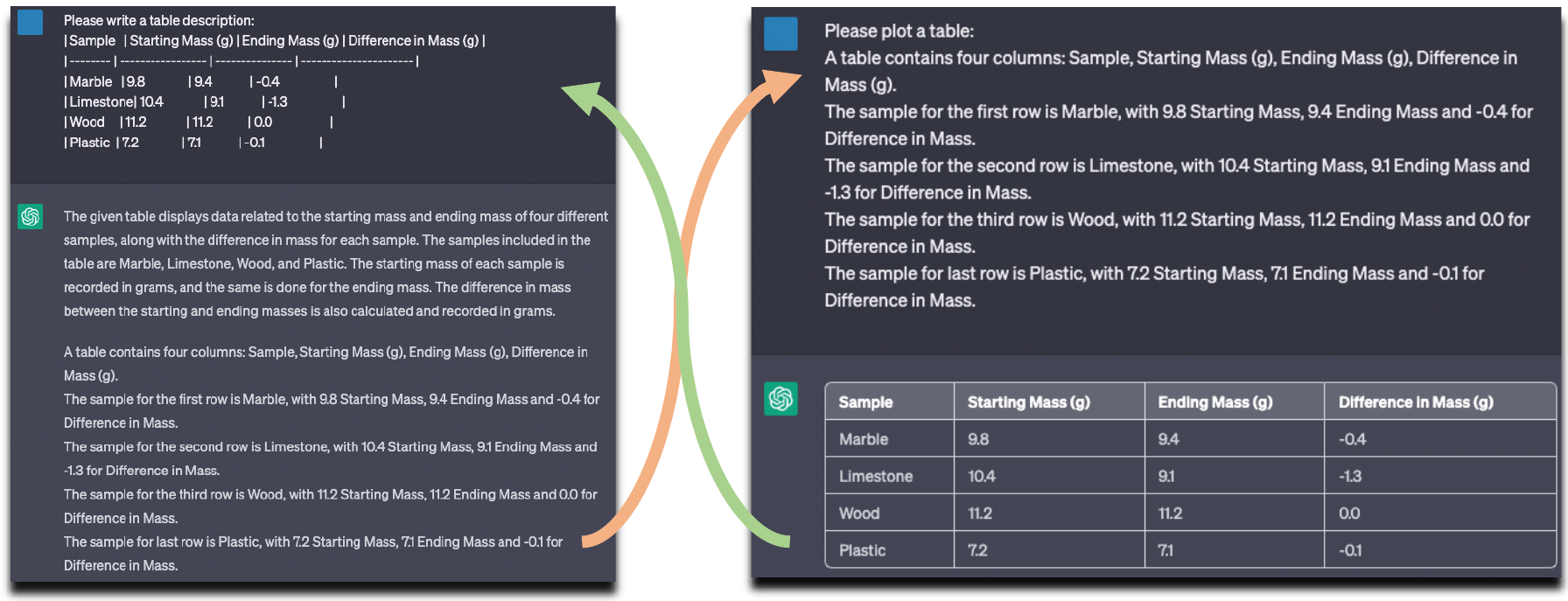}
  \caption{Demonstration of using ChatGPT for tabular data and table description transformation.}
  \label{fig:table_transform}
\end{figure}

Some questions in our dataset contain tabular data, which poses a challenge for smaller language models in terms of inputting and understanding structured data. To address this issue, as shown in Figure \ref{fig:table_transform}, we leverage ChatGPT's table-understanding capability to create table descriptions from the tabular data and verify the description correctness by having ChatGPT generate a table based on the description\footnote{Obtained with ChatGPT version 13th Feb 2023.}. Notably, we found that all the tabular data in our dataset could be accurately reconstructed based on the description generated by ChatGPT. Consequently, we replaced all the tabular data from the question part of our prompts with the corresponding generated descriptions.

\section{Further Experimental Details and Discussions} \label{app:experiments} 

\subsection{Experimental Setup} \label{app:experimental_setup}
\paragraph{Dataset}
In this paper, we employ the Hewlett Foundation: Short Answer Scoring (ASAP-SAS) dataset\footnote{\url{https://kaggle.com/competitions/asap-sas}}. This dataset encompasses over 23,000 short answer responses from students in grades 7 to 10, including ten questions spanning subjects such as Science, Biology, English, and Art. Expert human raters have manually scored each response on a scale of 0-2 or 0-3, based on predefined rubrics. Instead of focusing on assessment on the grammatical or writing side of the student responses, we are more interested in response assessment on STEM-related questions. Therefore, we only selected four subsets (\#1, \#2, \#5 and \#6) relating to Science and Biology from the ASAP-SAS datasets. We didn't include other subsets since they are either focused on English and Art or contain multi-modal data (e.g. Graphs) in the question that is difficult to be fed into language models. As the original dataset only provides the training and test sets, we created a development set by partitioning the training set in an 8:2 ratio. The detailed train, development, and test splits are shown in Table \ref{tab:dataset_stats}. 

\begin{table}[h!]
\centering
\small{
\begin{tabular}{@{}lcccc@{}}
\toprule
Subset & \textbf{\#1} & \textbf{\#2} & \textbf{\#5} & \textbf{\#6} \\ \midrule
% Subjects & Science & Science & Biology & Biology \\\midrule
% Score Scale &  \multicolumn{4}{c}{0-3} \\\midrule
\# Train & 1,338 & 1,023 & 1,436 & 1,438\\
\# Dev &334&255&359&359\\
\# Test &557&426&598 & 599\\\bottomrule
\end{tabular}}
\caption{Dataset statistics.}
\label{tab:dataset_stats}
\end{table}

\paragraph{Quadratic Weighted Kappa Implementation}
Quadratic Weighted Kappa, a widely used metric in evaluating the agreement between two raters in student response assessment, is defined as:
\begin{equation}
    \kappa = 1 - \frac{\sum_{i=1}^{k}\sum_{j=1}^{k} w_{ij} O_{ij}}{\sum_{i=1}^{k}\sum_{j=1}^{k} w_{ij} E_{ij}}
\end{equation}
where $k$ is the score set, $w$ is the weighted matrix, calculates as: $w_{i, j}=\frac{(i-j)^2}{(k-1)^2}$. $O$ is a $k\times k$ histogram matrix and $E$ being the $k\times k$ expected value matrix. 

\paragraph{Hyperparameter Settings}
We utilized the OpenAI API with the \texttt{gpt-3.5-turbo} model version 23 Mar 2023 for the generation of \textit{Simple}/\textit{Complex}/\textit{Example instruction}-based rationales. Default parameters were retained, with the temperature parameter set to 1.0. For our fine-tuning experiments, we deployed NVIDIA A100 80G graphics cards. The AERA fine-tuning procedure adopted the \texttt{Long-t5-tglobal-large} as the foundational model. Training for the rationale generation (RG) task was executed with a batch size of 8 over 30 epochs, while the text classification (TC) task used a batch size of 16 across the same number of epochs. We selected learning rates of 1e-5 for the TC task and 1e-4 for the RG task, implementing a weight decay of 0.01. To ensure robust performance metrics, each configuration was executed thrice for RG and five times for TC, using random seeds of 210, 102, 231, 314, and 146.

\paragraph{Model Implementation}
% We compare our method with two classification baselines: Bert-base-uncased model (BERT) fine-tuned with student answers as inputs only, Longformer-base-4096 (Longformer) with student answers as input only, and Longformer-base-4096 (Longformer-all) with Question, Key Elements, Rubric and Student Answers as input. 
We utilized the HuggingFace Transformer library\footnote{\href{https://github.com/huggingface/pytorch-transformers}{HuggingFace Transformer}} for the implementation of models such as BERT \cite{devlin2018bert}, Longformer \cite{Beltagy2020Longformer}, and LongT5\footnote{Given the extensive nature of the content within questions, key elements, and rubrics, the combined length with student responses typically exceeds 1,024 tokens. Consequently, our experiments employ models specifically designed to manage inputs from longer documents.} \cite{guo-etal-2022-longt5}.

\subsection{Faithfulness of ChatGPT-Generated Rationales w.r.t its Predicted Scores}

To the best of our knowledge, there is no established automated evaluation method for assessing the quality of ChatGPT-generated rationales. We proposed to design a proxy check to verify the faithfulness of the ChatGPT-generated rationale with respect to its predicted student answer assessment scores, which can be represented as $R\rightarrow Y$. We gathered the outputs produced by ChatGPT on our dataset and fine-tuned a text classifier to predict the score $\hat{y}_i$ using the generated rationale $\hat{r}_i$ as input. In this process, we did not perform any filtering. That is, some of the ChatGPT-predicted answer scores may be wrong. Our purpose is to establish a proxy check if the ChatGPT-generated rationales are indeed faithful explanations of its predicted answer scores.
% \begin{equation}
%     \hat{y}_i = p_{\rm LM}(\hat{r}_i)
% \end{equation}
As shown in Table \ref{tab:rationale_score_verify}, we observe a strong correlation between the ChatGPT-generated rationales and its predicted corresponding scores across all four datasets. % achieved high performance in accuracy. 
This finding suggests that the ChatGPT-generated rationales could be considered as somewhat faithful explanations of its predicted assessment scores. 
\begin{table}[h]
\centering
\small{
\begin{tabular}{@{}lcccc@{}}
\toprule
  & \textbf{\#1} & \textbf{\#2} & \textbf{\#5} & \textbf{\#6} \\ \midrule
Acc & 98.69 & 87.08 & 98.15 & 93.10\\
F1 & 98.77 & 83.47 & 97.18 & 95.47\\
QWK & 99.07 & 90.48 & 97.87 & 93.36\\\bottomrule
\end{tabular}}
\caption{Predictive performance on score classification output by ChatGPT using its generated rationales.}
\label{tab:rationale_score_verify}
\end{table}

\subsection{Simulatability of ChatGPT-Generated Rationales w.r.t its Predicted Scores}

\citet{wiegreffe-etal-2021-measuring} proposed a rationale quality evaluation method based on the association between generated rationale and the predicted label to evaluate the free-text rationale quality. Simulatability, instead of relying on the word-level overlap, assesses the ability of a generated rationale to predict the label by measuring the difference in task performance when the rationale is provided as input compared to when it is absent:
\begin{equation}
    acc(IR\rightarrow O) - acc(I\rightarrow O)
\end{equation}

We conducted an experiment to evaluate the simulatability of rationales generated by ChatGPT, as detailed in Table \ref{tab:sumulatability}. In this context, $XR\rightarrow Y$ denotes a generative classification setting fine-tuned on the Long T5 model. It takes into account questions, key elements, rubrics, student answers, and ChatGPT-generated rationales (using the Example Instruction template) as input, and outputs ChatGPT-predicted scores. Conversely, $X\rightarrow Y$ is tuned under the same classification setting but omits the rationale from the input.

Contrary to the consistency findings from \cite{wiegreffe-etal-2021-measuring}, where results trended toward 0, we noted positive disparities between $acc(XR\rightarrow Y)$ and $acc(X\rightarrow Y)$, as evident in the table's final row. This implies that rationales generated by ChatGPT, utilizing the \textit{Example Instruction} template, enhance label prediction, especially for datasets \#1, \#2, and \#5. While the accuracy difference for dataset \#6 is less than 0, there's a marked improvement in F1 and QWK metrics. This suggests that incorporating rationales into the input bolsters class sensitivity and aligns more closely with gold label scores.

In summary, across all datasets, the performance uptick indicates that ChatGPT-produced rationales exhibit commendable quality in simulatability tests. However, dataset \#6's outcomes hint that solely focusing on accuracy for evaluations might not be ideal for tasks with nuanced class sensitivity, such as student answer assessment.

\begin{table*}[ht]
\centering
\resizebox{\linewidth}{!}{
\begin{tabular}{@{}lccc|ccc|ccc|ccc@{}}
\toprule
\textbf{Dataset} (Subject)  & \multicolumn{3}{c}{\textbf{\#1} (Science)} & \multicolumn{3}{c}{\textbf{\#2} (Science)} & \multicolumn{3}{c}{\textbf{\#5} (Biology)} & \multicolumn{3}{c}{\textbf{\#6} (Biology)} \\
\textbf{Method/Model}  &  Acc & F1 & QWK &  Acc & F1 & QWK &  Acc & F1 & QWK &  Acc & F1 & QWK   \\\midrule 
$X\rightarrow Y$&69.96 &70.84 & 82.09 & 56.57 &53.12 &57.83 & 85.62 & 58.94 & 79.53& 89.20 & 62.86 &83.19\\
$XR\rightarrow Y$&80.91 &77.53 &85.70 & 82.39 & 80.52 & 87.61 &87.34 &82.56 &88.89  & 88.48 &76.83 &89.58 \\
$acc(XR\rightarrow Y) - acc(X\rightarrow Y)$& +10.95 & - & - & +25.82& - & - & +1.72 & - & - & -0.72 & - & - \\\bottomrule
\end{tabular}}
\caption{Analysis on ChatGPT-generated Rationales' Simulatability.}
% \ddag stands for trained with a train set with additional refined rationales and fixed incorrect labels.}
 \label{tab:sumulatability}
\end{table*}

\subsection{Results by Fine-Tuning Long T5 on Filtered ChatGPT Outputs} \label{app:filtered}

%\paragraph{Filtered Data Statistic}

In Table \ref{tab:filtered_dataset_stats}, we present the statistics of the training set after filtering out instances where ChatGPT predicts wrong answer scores. We observe that when using the \emph{Simple Instruction}, ChatGPT predicts correct answer scores for less than half of the instances. However, with the \emph{Complex Instruction}, there is a notable increase in the number of instances where ChatGPT outputs correct answer scores. Interestingly, the \emph{Example Instruction} does not yield improvements for dataset \#1, \#5, and \#6. But it enables ChatGPT to predict more correct answer scores for the dataset \#2.

%Although the amount of data is reduced for \#1, \#5, and \#6 with the Example Instruction prompted train set, the overall performance increases for the corresponding fine-tuned student models. This observation indicates that the amount of data is not necessarily correlated with performance when it is sufficient.

\begin{table}[h]
\centering
\small{
\begin{tabular}{@{}lcccc@{}}
\toprule
Subset & \textbf{\#1} & \textbf{\#2} & \textbf{\#5} & \textbf{\#6} \\ \midrule
\# Train & 1,338 & 1,023 & 1,436 & 1,438\\
Simple Instruction& 627 & 412 & 761 & 692\\
Complex Instruction&695&407&1,051&1,016\\
Example Instruction&689&477&968&987\\\bottomrule
\end{tabular}}
\caption{Statistics of the training set after filtering out incorrect ChatGPT-predicted answer scores.}
\label{tab:filtered_dataset_stats}
\end{table}

%\paragraph{Analysis}
In Table \ref{tab:model_comparison_filtered}, we show the results by fine-tuning Long T5 on the filtered ChatGPT outputs. 
Consistent with ChatGPT's inference performance, the fine-tuned %text generation model we trained using ChatGPT-generated rationales with correctly predicted scores 
Long T5 also exhibits performance improvement when trained on the filtered ChatGPT outputs produced using \emph{Complex} or \emph{Example Insutrctions} compared to \emph{Simple Instruction}. Interestingly, although the amount of data is reduced for subsets \#1, \#5, and \#6 when using the \emph{Example Instruction} compared to \emph{Complex Instruction} as shown in Table \ref{tab:filtered_dataset_stats}, the overall performance is the best for the  fine-tuned Long T5 models. We have conducted error analysis on the ChatGPT-generated outputs and found that the hallucination problem could be significantly reduced by providing demonstration examples in the \emph{Example Instruction} (More discussions can be found in \textsection{\ref{sec:analysis_hallucinations}}). For this reason, we decided to use the \emph{Example Instruction} in all our subsequent experiments.

%This observation indicates that the amount of data is not necessarily correlated with performance when it is sufficient.
%Interestingly, although outputs generated with \emph{Example Instruction} achieve the highest QWK and surpass ChatGPT's performance on most datasets (\#1, \#5, and \#6), fine-tuning on filtered correct ChatGPT-generated rationales can not guarantee its performance surpasses ChatGPT. Indicating the correct score filtering strategy may influence the quantity/quality of the data and result in a worse rationale generator. Since we observed a significant ChatGPT hallucination problem on zero-shot generated rationales (will be mentioned in \textsection{\ref{sec:analysis_hallucinations}}), to avoid and minimize the influence caused by hallucination, we focus on the investigation of Example Instruction generated rationales.

\begin{table*}[ht]
\centering
\resizebox{\linewidth}{!}{
\begin{tabular}{@{}lccc|ccc|ccc|ccc|ccc@{}}
\toprule
\textbf{Dataset} (Subject)  & \multicolumn{3}{c}{\textbf{\#1} (Science)} & \multicolumn{3}{c}{\textbf{\#2} (Science)} & \multicolumn{3}{c}{\textbf{\#5} (Biology)} & \multicolumn{3}{c}{\textbf{\#6} (Biology)} & \multicolumn{3}{c}{\textbf{Overall}} \\
\textbf{Method/Model}  &  Acc & F1 & QWK &  Acc & F1 & QWK &  Acc & F1 & QWK &  Acc & F1 & QWK &  Acc & F1 & QWK  \\\midrule 
Simple Instruction &43.39 & 32.39 & 40.01&23.71 & 9.97 & 0.69&68.56 & 32.82 & 45.29&79.69 & 37.14 & 64.95&53.84 & 28.08 & 37.74\\
Complex Instruction &47.16 & 38.36 & 54.48 & 40.61 & 29.3 & \textbf{38.3} & 79.21 & 42.14 & 61.63 & 85.70 & 43.26 & 67.73 & 63.17 & 38.27 & 55.54\\
Example Instruction & 56.79 & 55.95 & \textbf{69.96} & 35.60 & 24.02 & 23.94 & 78.76 & 48.79 & \textbf{71.83} & 84.19 & 54.93 & \textbf{79.17} & 63.84 & 45.92 & \textbf{61.23}\\\bottomrule
\end{tabular}}
\caption{Evaluating the performance of Long T5 models that have been fine-tuned using rationales generated by ChatGPT prompt with other templates.}
% \ddag stands for trained with a train set with additional refined rationales and fixed incorrect labels.}
 \label{tab:model_comparison_filtered}
\end{table*}

\subsection{Human Evaluation Details} \label{app:human_eval}
In this section, we provide further details and settings on our human evaluation experiments.

\subsubsection{Evaluation Setup}
\begin{figure*}[htbp]
  \centering
  \includegraphics[width=\linewidth]{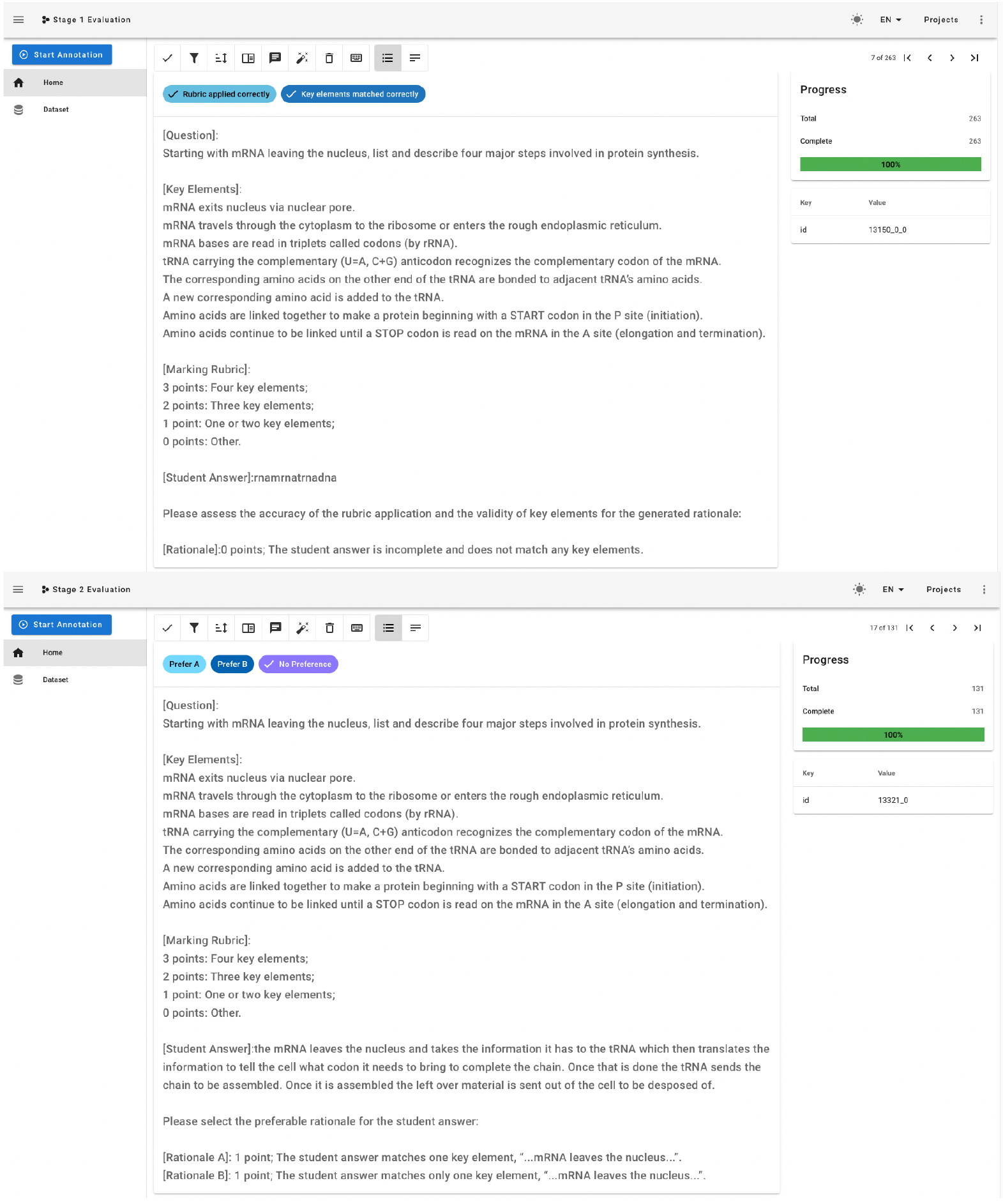}
  \caption{Screenshots of the annotation platform for both human evaluation tasks.}
  \label{fig:evaluation_plat}
\end{figure*}

\paragraph{Data Selection} We randomly selected 10\% of instances from the run with the highest QWK. Among the sampled data, we further selected 20\% for the purpose of calculating the Inter-Annotator Agreement (IAA) score. The detailed statistics of the total sampled data are shown in Table \ref{tab:human_eval_stats}.
\begin{table}[h]
\centering
\small{
\begin{tabular}{@{}lccccc@{}}
\toprule
Subset & \textbf{\#1} & \textbf{\#2} & \textbf{\#5} & \textbf{\#6}& \textbf{\#all} \\ \midrule
10\% Sampled &56&43&60&60&219\\
Duplicate for IAA &11&9&12&12&44\\
Total & 67&52&72&72&263\\\midrule
\multicolumn{5}{l}{Instances for Rationale Correctness}&526 \\
\multicolumn{5}{l}{Instances for Rationale Preference}&263 \\
\bottomrule
\end{tabular}}
\caption{The statistics of the sampled data for human evaluation.}
\label{tab:human_eval_stats}
\end{table}
\paragraph{Annotator} Two annotators are selected for the evaluation process. Both evaluators are PhD students with computer science backgrounds and have received training on the evaluation schema and the use of the annotation platform. Each assigned task took about 5 hours to complete, and the annotators were paid fairly at a rate of \$21.83/hour.
\paragraph{Evaluation Platform} As shown in Figure \ref{fig:evaluation_plat}, our evaluation is built with Docanno\footnote{\url{https://github.com/doccano/doccano}}. The labels are designed to indicate whether an option is considered correct or incorrect based on its selection or non-selection. %for selection as correct and incorrect for not selected.

\subsubsection{Human Evaluation Results}

\paragraph{IAA Results} We use Cohen Kappa for IAA analysis.
\[
\kappa = 1 - \frac{1 - P_o}{1 - P_e}
\]
where \(P_o\) is the relative observed agreement among raters (identical to accuracy), and \(P_e\) is the hypothetical probability of chance agreement, using the observed data to calculate the probabilities of each observer randomly seeing each category. Our IAA results in Table \ref{tab:iaa_results} show that the annotators exhibited moderate agreement on the correctness of key elements and the faithfulness of rubric, while they fairly agreed on the preference of rationales.

\begin{table}[h]
\centering
\small{
\begin{tabular}{@{}lc@{}}
\toprule
\textbf{Tasks} & \textbf{IAA Score}\\ \midrule
Correctness of Key Elements & 0.4579\\
Faithfulness of Rubric & 0.5056\\
Rationale Preference & 0.3276\\
\bottomrule
\end{tabular}}
\caption{Inter-Annotator Agreement results.}
\label{tab:iaa_results}
\end{table}

\paragraph{More Detailed Evaluation Results} We present a breakdown of evaluation results for both human evaluation tasks, showing the percentage of correctness selections by each annotator in Table \ref{tab:stage1}, and by each subset in Table \ref{tab:break_down_by_set}.
\begin{table}[h]
\centering
\small{
\begin{tabular}{@{}lccc@{}}
\toprule
 & Ann 1 & Ann 2 & Total\\ \midrule
\multicolumn{4}{c}{Human Evaluation on Rationale Correctness}\\ \midrule
Key Elements on AERA   & 0.86 & 0.80 & \textbf{0.83}\\
Rubric on AERA         & 0.96 & 0.92 & \textbf{0.94}\\
Key Elements on ChatGPT& 0.60 & 0.38 & 0.52\\
Rubric on ChatGPT      & 0.86 & 0.87 & 0.86\\\midrule
\multicolumn{4}{c}{Human Evaluation on Rationale Preference}\\ \midrule
Prefer AERA& 0.57 &0.50 & \textbf{0.54}\\
Prefer ChatGPT& 0.24 & 0.22 & 0.23\\
No Preference& 0.17 & 0.28  & 0.23\\
\bottomrule
\end{tabular}}
\caption{A breakdown of evaluation results by each annotator.}
\label{tab:stage1}
\end{table}

\begin{table}[h]
\centering
\small{
\begin{tabular}{@{}lcccc@{}}
\toprule
 & \#1 & \#2 & \#5 & \#6\\ \midrule
\multicolumn{5}{c}{Human Evaluation on Rationale Correctness}\\ \midrule
Key Elements on AERA   & 0.92 & 0.72 & 0.90 & 0.73\\
Rubric on AERA         & 1.00 & 0.78 & 1.00 & 0.93\\
Key Elements on ChatGPT& 0.48 & 0.51 & 0.53 & 0.54\\
Rubric on ChatGPT      & 0.91 & 0.85 & 0.96 & 0.73\\\midrule
\multicolumn{5}{c}{Human Evaluation on Rationale Preference}\\ \midrule
Prefer AERA& 0.56 &0.48 & 0.56 & 0.53\\
Prefer ChatGPT& 0.16  & 0.29  & 0.24 & 0.26\\
No Preference& 0.27 & 0.23 & 0.21 & 0.21\\
\bottomrule
\end{tabular}}
\caption{A breakdown of evaluation results by each subset.}
\label{tab:break_down_by_set}
\end{table}

\subsection{Analysis of ChatGPT Hallucinations} \label{sec:analysis_hallucinations}
In this section, we discuss various hallucination %\footnote{We adopt the definition from \citet{bang2023multitask}, where hallucination stands for model-generated statements from its parametric memory that cannot be verified from the source.} 
cases observed in the ChatGPT-generated rationales under the zero-shot setting, i.e., using either \emph{Simple Instruction} or \emph{Complex Instruction} as described in \textsection{\ref{sec:zero-shot-rationale}} and without supplying any demonstration examples. %, . 
Table \ref{tab:chatgpt_hallucination} demonstrates cases of inconsistent and inaccurate assessments, which can be grouped into five types: \textbf{(1) Incorrect scoring scale}. Despite providing a clear 0-3 integer score rubric, ChatGPT occasionally generates rationales that include incorrect score caps, such as 5 or 12, or even fractional score scales possibly stem from its knowledge base. \textbf{(2) Inconsistent assessment}.  Some rationales display completely contradictory scores in two different places in the rationale text. \textbf{(3) Uncertain score prediction}. %Similar to the first case, zero-shot rationales 
In some cases, ChatGPT may ignore the marking rubric and outputs %may disregard the rubric, resulting in 
uncertain scores such as `\emph{1-2 points}'. \textbf{(4) Factual mistake}.  We observed instances where the matched key elements identified in the generated rationales %notice some rationale declared correctly matched key elements 
were never mentioned by the student's answer or included in the key answer elements. \textbf{(5) Vague rationale}. We observed that zero-shot generated rationales often provide vague or irrelevant explanations for  student response, which may not be helpful for feedback and could be difficult to understand.

In contrast, using the \emph{Example Instruction} prompt by supplying some demonstration examples guides ChatGPT to follow a structured format for rationale generation and answer scoring. %few-shot examples showcase the rationale generation style and scoring, leading to a more structured format. 
Moreover, instructions that are oriented towards examples help ChatGPT to rely less on its knowledge base and instead utilise information from the provided resources such as key answer elements and marking rubric. Our analysis reveals that the hallucination problem can be partly alleviated by using the \emph{Example Instruction} prompt with demonstration examples. Consequently, we have chosen the \emph{Example Instruction} prompt %example-oriented few-shot rationale generation 
as our primary rationale generation method.

\begin{table}[ht!]
  \centering
  % \resizebox{0.9\linewidth}{!}

  \begin{tabular}{p{0.97\linewidth}}
    \toprule
    \textbf{Incorrect scoring scale:} \\
    ... answer should receive \myorange{1 point out of 5}. \\
    ... answer should receive \myorange{1.5 points out of 3}. \\
    ... Overall, this student answer receives \myorange{a score of 2 out of 12 (0+0+1+1)} as the answer does not accurately and completely ...\\
    \midrule
    \textbf{Inconsistent assessment:} \\
    \myorange{Score: 1 point} This student answer ... Therefore, the answer is not relevant to the question and should \myorange{receive a score of 0 points}.\\
    \midrule
    \textbf{Uncertain score prediction:} \\
    ... Therefore, \myorange{this answer would receive a score of 1-2 points out of 3}.\\
    \midrule
    \textbf{Factual mistake:} \\
    ... this Student answer includes three of the key elements: \myorange{selective permeability}, passive transport, and facilitated diffusion\\
    \midrule
    \textbf{Vague rationale:} \\
    ... the answer demonstrates some understanding of protein synthesis but is missing several key elements and contains some inaccuracies.\\
    \bottomrule
  \end{tabular}
  \caption{ChatGPT hallucination examples from the rationales generated using either \emph{Simple} or \emph{Complex} Instruction under the zero-shot setting.}
  \label{tab:chatgpt_hallucination}
\end{table}

\subsection{Example Rationales Generated using AERA vs. ChatGPT}

%In this section, we present an analysis of the student model assessment results and generated rationales, comparing them with the teacher model ChatGPT's outcomes, as illustrated in 
Table \ref{tab:rationale_gen_examples} shows example rationales generated using the student model, Long T5, in comparison with those generated by the teacher model, ChatGPT. We observe that both the Long T5- and ChatGPT-generated results follow the same structured format as demonstrated in the examples provided in the prompt, that a score is given first, followed by a rationale explaining the scoring decision.
%exhibit similarities to ChatGPT's results, with both presenting clear scores followed by well-defined rationales that highlight the reasoning behind the scores.

The refinement of the training data, which involved cleaning and correcting some inaccurately generated rationales by providing the actual answer scores as input to ChatGPT, has led to a stronger correlation between Long T5-generated rationales and the predicted scores. On the contrary, the ChatGPT-generated results for \#1, \#2, and \#6 exhibit minor discrepancies due to over-matching or under-matching certain key elements.

We also noticed a small number of mistakes in the Long T5-generated results, primarily attributable to the students' vague descriptions, making it difficult for the language model to compare the answers with the key elements. Additionally, some questions include rubrics such as "\textit{other acceptable responses}", which are particularly challenging for language models to assess, given their lack of domain-specific background knowledge.

In summary, our distilled Long T5 model demonstrates a strong capability to assess student responses and generate accurate rationales. Despite the occasional errors and challenges posed by vague student answers and certain rubrics, the model's overall performance is promising for applications in educational settings.

\subsection{Explore the Influence of Number of Demonstration Examples}
In this section, we have performed an ablation study on the influence of test performance by the number of demonstrations provided to ChatGPT\footnote{ChatGPT version 3 Aug 2023.}. In this experiment, we gradually reduced the number of demonstration examples included in the prompt to find out the influence on the performance. As we present on the \#6 in Table \ref{tab:chatgpt_performance}, aligned with observations reported in prior work \citep{brown2020language}, the test performance achieves the highest with all the demonstration examples included.

\subsection{Investigate the Generalizability of AERA}
We wanted to demonstrate that our approach is applicable in a wide range of scenarios. To do this, we conducted an ablation study called "leave one out", training our framework on three subsets and testing it on the left subset. The results, as shown in Table \ref{tab:subset_performance}, indicate that our framework can not only evaluate student answers based on the trained question, key elements and rubric; but also generalize well beyond to unseen datasets.

\subsection{Rationale Generation from Other LLMs}
This section presents an example from the \#5 to demonstrate that our prompting strategy is still effective for models other than ChatGPT, such as Bard or FlanT5. During the experiment design phase, we primarily focused on ChatGPT due to its robust capabilities and cost-effectiveness. The largest open-sourced model We experimented with was the LLaMA-2 70B. However, as shown in Table \ref{tab:other_llm}, the model struggled to produce coherent rationales and often repeated the marking rubrics in its response.

\begin{table}[ht]
\centering
\resizebox{\linewidth}{!}{
\begin{tabular}{clcccc}
\toprule
Demo amount & Scores & Acc & F1 & QWK \\
\midrule
5 & 0,1,2,3,3 & \textbf{79.97} & \textbf{54.66} & \textbf{71.48} \\
4 & 0,1,2,3 & 77.96 & 40.75 & 68.63 \\
3 & 0,1,2 & 74.62 & 48.75 & 66.24 \\
2 & 0,1 & 72.62 & 42.82 & 54.90 \\
1 & 0 & 67.45 & 39.30 & 57.26 \\
\bottomrule
\end{tabular}}
\caption{Ablation study on the number of demonstrations on \#6.}
\label{tab:chatgpt_performance}
\end{table}

\begin{table}[ht]
\centering
\resizebox{\linewidth}{!}{
\begin{tabular}{@{}ccccc@{}}
\toprule
Train Sets & Test Sets & Acc & F1 & QWK \\
\midrule
\#2, \#5, \#6 & \#1 & 57.99 & 56.05 & 71.20 \\
\#1, \#5, \#6 & \#2 & 44.84 & 40.34 & 58.42 \\
\#1, \#2, \#6 & \#5 & 84.95 & 39.49 & 63.02 \\
\#1, \#2, \#5 & \#6 & 85.14 & 54.10 & 75.21 \\
\bottomrule
\end{tabular}}
\caption{"leave one out" experiment for method generalizability.}
\label{tab:subset_performance}
\end{table}

\begin{table*}[ht!]
  \centering
  \resizebox{\linewidth}{!}{
  \begin{tabular}{p{\linewidth}}
    \toprule
    \textbf{\#1 Student Answer:} To replicate the group's experiment, the procedure would have to state how much vinegar was poured into the containers. Also, they should specify what kinds of containers they were using as this could affect the results. In the procedure, they also failed to tell what 4 samples were used in the experiment, a key variable to it's success. \textbf{Score: 3}\\
    \textbf{Long T5:} 3 points; This response describes three additional pieces of information that would be needed to accurately replicate the experiment: ‚"how much vinegar was poured into the containers...what kinds of containers they were using...and what 4 samples were used in the experiment.‚" \\ 
    \textbf{ChatGPT:} 2 points; This response describes two additional pieces of information that would be needed to accurately replicate the experiment: ‚"how much vinegar was poured into the containers‚" and ‚"what 4 samples were used in the experiment.‚" It also hints at a potential third piece of information needed: ‚"what kinds of containers were used.‚" \\ 
    \midrule
    \textbf{\#2 Student Answer:} a. Based on the data, the strectchiest polymer plastic of the four is plastic type B, because in the both trials, it stretched the most (22 and 23 mm). b. 1.) Add more trials to increase accuracy of the results. 2.) Repeat the experiment with heavier weights to see if it affects the results. \textbf{Score: 2}\\
    \textbf{Long T5:} 2 points; The student provides an acceptable conclusion based on the data: ‚"...the stretchiest polymer plastic of the four is plastic type B....‚" One correct way to improve the experimental design and/or the validity of the results is given: ‚"Add more trials to increase accuracy of the results.‚" The second improvement given, ‚"...trying the experiment with heavier weights to see if it affects the results,‚" is too vague and does not receive credit. \\ 
    \textbf{ChatGPT:} 3 points; The student provides an acceptable conclusion based on the data: ‚"...the stretchiest polymer plastic of the four is plastic type B...‚" Two correct ways to improve the experimental design and/or the validity of the results are given: ‚"1.) Add more trials to increase accuracy of the results....‚" and ‚"2.) Repeat the experiment with heavier weights to see if it affects the results.‚"\\ 
    \midrule
    \textbf{\#5 Student Answer:} tRNA will transfer it out. \textbf{Score: 0}\\
    \textbf{Long T5:} 0 points; The student answer does not match any key elements given.\\ 
    \textbf{ChatGPT:} 0 points; The student answer does not provide any coherent or relevant information on the steps involved in protein synthesis.\\ 
    \midrule
    \textbf{\#6 Student Answer:} Osmosis, Is the movement of water across the membrain. Endocitosis, is the movement of things in to the cell. Exocitosis, is the movement of things out of the cell. \textbf{Score: 1}\\
    \textbf{Long T5:} 1 point; This student answer only matches one key element, "Osmosis... movement of water". The other two concepts are incorrect or incomplete.\\ 
    \textbf{ChatGPT:} 2 points; This student answer matches two key elements, "Osmosis... movement of water across the membrane" and "Endocytosis... movement of things into the cell" but didn't include an explanation for "Exocytosis".\\ 
    \bottomrule
  \end{tabular}}
  \caption{Examples of AERA generated rationales compared with ChatGPT results.}
  \label{tab:rationale_gen_examples}
\end{table*}

\begin{table*}[ht]
\centering
\resizebox{\linewidth}{!}{
\begin{tabular}{p{18cm}}
\toprule
\textbf{\textit{Given this student answer from \#5 with the Example Instruction prompt template}}:\\ ``1. tRNA tranfers it outside the nucleus to be matched with a ribosome2. The ribosome latches on to an amino acid.3. The codons match up4. It creates protein.''\\
\midrule
\textbf{\textit{ChatGPT}}: 1 point; The student answer only matches one key element, “...tRNA transfers it outside the nucleus to be matched with a ribosome...”. \\
\midrule
\textbf{\textit{GPT-4}}: 1 point; The student answer matches two key elements, "The codons match up" and "It creates protein". \\
\midrule
\textbf{\textit{Google Bard:}} The student answer matches only two key elements, ‘tRNA tranfers it outside the nucleus to be matched with a ribosome’ and ‘The codons match up’, so the score is 1 point. Here are the four major steps involved in protein synthesis, starting with mRNA leaving the nucleus: Initiation: mRNA leaves the nucleus and binds to a ribosome in the cytoplasm. The ribosome then scans the mRNA for a start codon (AUG). Elongation: A tRNA molecule with an amino acid attached to it binds to the ribosome at the A site. The anticodon of the tRNA matches the start codon on the mRNA. The amino acid is then added to the growing polypeptide chain. The ribosome then moves one codon down the mRNA. Termination: The ribosome continues to read the mRNA until it reaches a stop codon. A release factor binds to the stop codon and the polypeptide chain is released from the ribosome. Folding and modification: The newly synthesized polypeptide chain is folded into its correct shape. It may also be modified by the addition of chemical groups. The student answer does not mention the initiation step, which is a critical step in protein synthesis. Therefore, the answer is not complete and the score is 1 point. \\
\midrule
\textbf{\textit{Flan T5 XXL}}: \textit{(Exceeded max input sequence length)} 1 point; The student answer matches one key element, “...tRNA tranfers it outside the nucleus to be matched with a ribosome2. The ribosome latches on to an amino acid3. The codons match up4. It creates protein.”. \\
\midrule
\textbf{\textit{LLaMA 2 70B Chat HF}}: Sure, here's the scoring rubric for the student answers: [Student Answer]: 1. tRNA tranfers it outside the nucleus to be matched with a ribosome2. The ribosome latches on to an amino acid.3. The codons match up4. It creates protein. [Rationale and Score]: The student answer matches two key elements: "tRNA tranfers it outside the nucleus to be matched with a ribosome" (step 1) and "It creates protein" (step 4).* However, the student answer does not mention the other two key elements: "mRNA bases are read in triplets called codons (by rRNA)" (step 2) and "Amino acids are linked together to make a protein beginning with a START codon in the P site (initiation)" (step 3). Score: 2 points Here's the corrected scoring rubric:* 4 points: Four key elements;* 3 points: Three key elements;* 2 points: Two key elements;* 1 point: One key element;* 0 points: Other. \\
\bottomrule
\end{tabular}}
\caption{Example output for answer assessment and rationale generation from other popular LLMs.}
\label{tab:other_llm}
\end{table*}

\newpage
\section{Prompt Details} \label{app:prompt_detail}
In this section, we provide the full detail of the question, key elements and rubric we used in the prompt for each dataset. We highlighted table descriptions generated via Tabular Data Transformation mentioned in \textsection{\ref{app:tabular_transform}} \myorange{in orange}.
\subsection{Subset \#1}
[Question]: \\
A group of students wrote the following procedure for their investigation. \\
Procedure: \\
1.Determine the mass of four different samples. \\
2.Pour vinegar in each of four separate, but identical, containers. \\
3.Place a sample of one material into one container and label. Repeat with remaining samples, placing a single sample into a single container. \\
4.After 24 hours, remove the samples from the containers and rinse each sample with distilled water. \\
5.Allow the samples to sit and dry for 30 minutes. \\
6.Determine the mass of each sample. \\
\myorange{The students's data are recorded in the table below. \\
A table contains four columns: Sample, Starting Mass (g), Ending Mass (g), Difference in Mass (g). \\
The sample for the first row is Marble, with 9.8 Starting Mass, 9.4 Ending Mass and -0.4 for Difference in Mass. \\
The sample for the second row is Limestone, with 10.4 Starting Mass, 9.1 Ending Mass and -1.3 for Difference in Mass. \\
The sample for the third row is Wood, with 11.2 Starting Mass, 11.2 Ending Mass and 0.0 for Difference in Mass. \\
The sample for last row is Plastic, with 7.2 Starting Mass, 7.1 Ending Mass and -0.1 for Difference in Mass.} \\
After reading the group's procedure, describe what additional information you would need in order to replicate the experiment. \\
Make sure to include at least three pieces of information. \\

\noindent[Key Elements]: \\
Needed Information: \\
You need to know how much vinegar was used in each container. \\
You need to know what type of vinegar was used in each container. \\
You need to know what materials to test. \\
You need to know what size/surface area of materials should be used. \\
You need to know how long each sample was rinsed in distilled water. \\
You need to know what drying method to use. \\
You need to know what size/type of container to use. \\
Other acceptable responses. \\

\noindent[Rubric]: \\
3 points: The response describes three additional pieces of information that would be needed to accurately replicate the experiment; \\
2 points: The response describes two additional pieces of information that would be needed to accurately replicate the experiment; \\
1 point: The response describes one additional piece of information that would be needed to accurately replicate the experiment; \\
0 point: The response describes little or no accurate or relevant information from the acid rain investigation. 

\subsection{Subset \#2}
[Question]:  \\
A student performed the following investigation to test four different polymer plastics for stretchability. \\
Procedure: \\
1. Take a sample of one type of plastic, and measure its length. \\
2. Tape the top edge of the plastic sample to a table so that it is hanging freely down the side of the table. \\
3. Attach a clamp to the bottom edge of the plastic sample. \\
4. Add weights to the clamp and allow them to hang for five minutes. \\
5. Remove the weights and clamp, and measure the length of the plastic types. \\
6. Repeat the procedure exactly for the remaining three plastic samples. \\
7. Perform a second trial (T2) exactly like the first trial (T1). \\
The student recorded the following data from the investigation. \\
\myorange{The table shows the amount of stretch (in millimeters) for four different types of plastic, labeled as A, B, C, and D, when subjected to two different stretching forces, labeled as T1 and T2. \\
For plastic type A, it stretched 10mm under T1 and 12mm under T2. \\
For plastic type B, it stretched 22mm under T1 and 23mm under T2. \\
For plastic type C, it stretched 14mm under T1 and 13mm under T2. \\
Lastly, for plastic type D, it stretched 20mm under both T1 and T2.} \\
a. Draw a conclusion based on the student’s data. \\
b. Describe two ways the student could have improved the experimental design and/or validity of the results. \\

\noindent[Key Elements]: \\
Conclusions: \\
Plastic sample B has more stretchability than the other polymer plastics. \\
Plastic sample A has the least amount of stretchability compared to the other polymer plastics. \\
Not all polymer plastics have the same stretchability. \\
Different polymer plastics have different stretchability (and are therefore suited for different applications). \\
A reasonable conclusion cannot be drawn due to procedural errors. \\
Other reasonable conclusions  \\
Experimental Design Improvements: \\
Provide the before and after measurements for length (Did the samples all start out the same size?). \\
Make sure the samples are all of the same thickness. \\
Variations in thickness could have caused variations in stretchability. \\
Perform additional trials. \\
Some of the samples have similar stretchability (A and C, B and D). \\
Two trials may not be enough to conclusively state that one is more stretchable than the other. \\
Indicate how many weights were added to the clamps (Was it the same number for each sample?). \\
Other acceptable responses \\

\noindent[Rubric]: \\
3 points: The response draws a valid conclusion supported by the student’s data and describes two ways the student could have improved the experimental design and/or the validity of the results; \\
2 points: The response draws a valid conclusion supported by the student’s data and describes one way the student could have improved the experimental design and/or the validity of the results. -or- The response describes two ways  \\the student could have improved the experimental design and/or the validity of the results but fails to draw or incorrectly draws a conclusion from the student’s data; \\
1 point: The response draws a valid conclusion supported by the student’s data but fails to describe, or incorrectly describes, how the student could have improved the experimental design and/or the validity of the results. -or-  \\The response describes one way the student could have improved the experimental design and/or the validity of the results but fails to draw or incorrectly draws a conclusion from the student's data.; \\
0 points: The response provides little or no correct information from the polymer investigation. 

\subsection{Subset \#5}
[Question]: \\
Starting with mRNA leaving the nucleus, list and describe four major steps involved in protein synthesis. \\

\noindent[Key Elements]:\\
mRNA exits nucleus via nuclear pore.\\
mRNA travels through the cytoplasm to the ribosome or enters the rough endoplasmic reticulum.\\
mRNA bases are read in triplets called codons (by rRNA).\\
tRNA carrying the complementary (U=A, C+G) anticodon recognizes the complementary codon of the mRNA.\\
The corresponding amino acids on the other end of the tRNA are bonded to adjacent tRNA’s amino acids.\\
A new corresponding amino acid is added to the tRNA.\\
Amino acids are linked together to make a protein beginning with a START codon in the P site (initiation).\\
Amino acids continue to be linked until a STOP codon is read on the mRNA in the A site (elongation and termination). \\

\noindent[Rubric]:\\
3 points: Four key elements;\\
2 points: Three key elements;\\
1 point: One or two key elements;\\
0 points: Other.

\subsection{Subset \#6}
[Question]: \\
List and describe three processes used by cells to control the movement of substances across the cell membrane.\\

\noindent[Key elements]: \\
Selective permeability is used by the cell membrane to allow certain substances to move across.\\
Passive transport occurs when substances move from an area of higher concentration to an area of lower concentration.\\
Osmosis is the diffusion of water across the cell membrane.\\
Facilitated diffusion occurs when the membrane controls the pathway for a particle to enter or leave a cell.\\
Active transport occurs when a cell uses energy to move a substance across the cell membrane, and/or a substance moves from an area of low to high concentration, or against the concentration gradient. \\
Pumps are used to move charged particles like sodium and potassium ions through membranes using energy and carrier proteins.\\
Membrane-assisted transport occurs when the membrane of the vesicle fuses with the cell membrane forcing large molecules out of the cell as in exocytosis.\\
Membrane-assisted transport occurs when molecules are engulfed by the cell membrane as in endocytosis.\\
Membrane-assisted transport occurs when vesicles are formed around large molecules as in phagocytosis.\\
Membrane-assisted transport occurs when vesicles are formed around liquid droplets as in pinocytosis.\\
Protein channels or channel proteins allow for the movement of specific molecules or substances into or out of the cell.\\

\noindent[Rubric]:\\
3 points: Three key elements;\\
2 points: Two key elements;\\
1 point: One key element;\\
0 points: Other.
\end{document}